%% file: acl2023.tex
\pdfoutput=1

\documentclass[11pt]{article}

\usepackage[]{ACL2023}

\usepackage{times}
\usepackage{latexsym}

\usepackage[T1]{fontenc}

\usepackage[utf8]{inputenc}
\usepackage{graphicx}
\usepackage{microtype}

\usepackage{booktabs}
\usepackage{makecell}
\usepackage{array}
\usepackage{multirow}
\usepackage{mathrsfs}

\usepackage{inconsolata}

\newcommand{\ours}{\textbf{\textsc{EG\textsuperscript{3}P}}}

\title{Explanation Graph Generation via Generative Pre-training over Synthetic Graphs}

\author{
        Han Cui, Shangzhan Li,  Yu Zhang\thanks{ \ \    Corresponding author.} \and Qi Shi \\
        Research Center for Social Computing and Information Retrieval, \\ Harbin Institute of Technology,   Harbin, China\\
        \{hcui, szli, zhangyu, qshi\}@ir.hit.edu.cn}

\begin{document}
\maketitle
\begin{abstract}
The generation of explanation graphs is a significant task that aims to produce explanation graphs in response to user input, revealing the internal reasoning process. This task is challenging due to the significant discrepancy between unstructured user queries and structured explanation graphs. 
Current research commonly fine-tunes a text-based pre-trained language model on a small downstream dataset that is annotated with labeled graphs. However, due to the limited scale of available datasets, this approach may prove to be insufficient in bridging the gap between natural language text and structured graphs.
In this paper, to alleviate the above limitations, we propose a novel pre-trained framework \ours (for \textbf{E}xplanation \textbf{G}raph \textbf{G}eneration via \textbf{G}enerative \textbf{P}re-training over synthetic graphs)\ for the explanation graph generation task.
Specifically, we first propose a text-to-graph generative task to pre-train the model with the goal of bridging the text-graph gap. Additionally, we propose an automatic corpus synthesis strategy for synthesizing a large scale of high-quality corpus, reducing the reliance on costly manual annotation methods.
Experimental results on ExplaGraphs show the effectiveness of \ours \ that our model surpasses all baseline systems with remarkable margins.
Besides, further analysis demonstrates that \ours \ is able to generate better explanation graphs on actual reasoning tasks such as CommonsenseQA and OpenbookQA.\footnote{Our code, checkpoints and corpus is released in \href{https://github.com/cccccent/EG3P}{https://github.com/cccccent/EG3P}}

\end{abstract}

\input{sections/introduction.tex}

\input{sections/Background_overview.tex}

\input{sections/2_The_Text2graph_Pre-training_Strategy.tex}

\input{sections/3_The_Construction_of_Synthetic_Corpus.tex}

\input{sections/4_experiments.tex}

\input{sections/5_results_and_analysis.tex}

\input{sections/related_work.tex}

\section{Conclusion}
In this paper, we propose a pre-training framework \ours\ for a structured explanation generation task. 
Distinct from existing pre-training tasks based on natural language text, \ours\ focuses more on training mapping between natural language and graphs. 
Meanwhile, due to the high cost of manually tagging, we construct queries from the synthetic graph automatically to get a large-scale corpus to support the pre-training process.
Using ExplaGraph as a main benchmark, experimental results show that \ours\ could significantly improve the ability of the model to generate explanations. 
In addition, on the other dataset, the results of the model after pre-training also showed a considerable improvement. Our approach offers a new possibility for addressing the challenges of limited labeled data in natural language processing tasks.

In the future, the ability of the model to generate explanation graphs will benefit from more datasets released with labels and more and more objective evaluation indicators put forward. 
Additionally, while our current approach processes graphs as strings, utilizing a model architecture that is more suitable for graph generation may further enhance the model's graph generation ability.

\section*{Limitations}
In our experiments, the most significant limitation is the lack of computational resources. 
Experimental results in this paper and previous work\cite{saha-etal-2022-explanation} have shown that a larger scale of  models could lead to higher structural and semantic accuracy of explanation graphs in this task. 
Constrained by computational resources, BART-Large is the largest model on which we can perform the complete process of experiments. We believe that graph generation would be better if sufficient resources were available to perform synthetic data based pre-training on a larger model.
In addition, since the evaluation metrics for graph generation tasks are incomplete yet, we can only evaluate a few samples manually outside of the metrics of the dataset, which is more subjective. 
With more evaluation methods with standardized processes proposed, the results of the experiment will be evaluated more objectively.

\section*{Acknowledgements}
We would like to thank the anonymous reviewers for their helpful comments. This work was supported by the National Natural Science Foundation of China (No.61976068 and No.62277002).

\section*{Ethics Statement}
In this paper, we propose a pre-training framework based on synthetic data to improve the ability of the model to generate explanation graphs. 
The datasets and model we used are all open-source and all the references that draw on the work of others are marked with citations. 
In the process of constructing the corpus, we ensure that all the triples come from ConceptNet, an open source knowledge base. 
All the steps involving selection are completely random, so that nothing such as bias or discrimination is introduced in following steps. 
Finally, our approach is designed to improve the interpretability of the model and won’t deviate from the semantics in the input text, so there is no ethical issues in this work.

\bibliography{anthology,custom}
\bibliographystyle{acl_natbib}

\newpage
\appendix

\section{The Distribution of Relations in the Synthetic Corpus}
\label{app:relationDistribution}
Table~\ref{tab:relationDistribution} shows the distribution of the 16 relations in the synthetic data. In the process of corpus construction, the process of relation extraction is completely random, so the distribution of relations in corpora of different sizes remains consistent. The only difference between the groups of experiments is the size of the corpus.

\section{Samples of Input and Output}
\label{app:inputexampleappendix}
Table~\ref{tab:inputSample} shows the samples from the synthetic data and different datasets. The samples shown in the upper part of the table are from the pre-training task, and the samples shown in the lower part are from the generation process of the downstream dataset.

\section{The Templates For Converting Triples to Natural Language}
\label{app:templates}
Table~\ref{tab:templates} shows all the templates we use in the experiments. We construct several different templates for a single relation to ensure the diversity after converting and randomly select one template each time for the expression of the triple. 

\section{The Metrics in ExplaGraphs}
\label{app:metrics}
The evaluation part of the main experiment directly adopts the evaluation metrics in the ExplaGraphs dataset. For a generated graph of the model, it must be ensured that the result of stance prediction of the corresponding instance is correct before entering the subsequent evaluation steps. The metrics of the graph mainly include the following points:

\paragraph{Structural Correctness Accuracy of Graphs (StCA)}
The StCA metric represents the proportion of all generated graphs that satisfy the structural constraints introduced following:
\begin{itemize}
    \item[*] The explanation graph needs to contain at least two nodes from the belief and two nodes from the argument. Each node has three words at most. The relationships represented by the edges in the graph must all come from predefined relationships.
    \item[*] The explanation graph must be a connected DAG.
    \item[*] The number of triples in the explanation graph must be between 3 and 8.
\end{itemize}

\paragraph{Semantic Correctness Accuracy of Graphs (SeCA)}
The SeCA metric judges the relationship (support, counter or incorrect) between the belief and the generated graph. If the judgment is consistent with the results of the stance prediction task, we consider the graph semantically correct. A RoBERTa-based three-class model is used during this process, and the evaluation process for this metric is entirely independent of structural features.

\paragraph{G-BERTScore (G-BS)}
This metric is based on BERTScore\cite{https://doi.org/10.48550/arxiv.1904.09675} and calculates the similarity between standard graph and generated one. We treat the graph as a set of edges and try to find the best matching result between the predicted graph and the gold one, where the similarity between the two edges is calculated according to the F1 value of BERTScore given the best assignment.

\paragraph{Graph Edit Distance (GED)}
The edit distance of a graph measures the difference between the generated graph and the gold graph, specifically referring to the number of changing operations that need to be performed between the two, where the changing operation refers to adding, deleting, or replacing edges or nodes in the graph.

\paragraph{Edge Importance Accuracy (EA)}
Edge importance accuracy measures the proportion of edges that are important enough in the generated graph. We consider an edge important as long as removing this edge will lead to a significant drop in the accuracy of stance prediction. This part of the evaluation is based on a separate RoBERTa model. 

\section{Case Study}
\label{app:caseStudy}
\subsection{Case study on ExplaGraphs}
\label{app:caseStudyExplaGraphs}
\begin{figure*}[t]
        \centering
        \includegraphics[width=14cm]{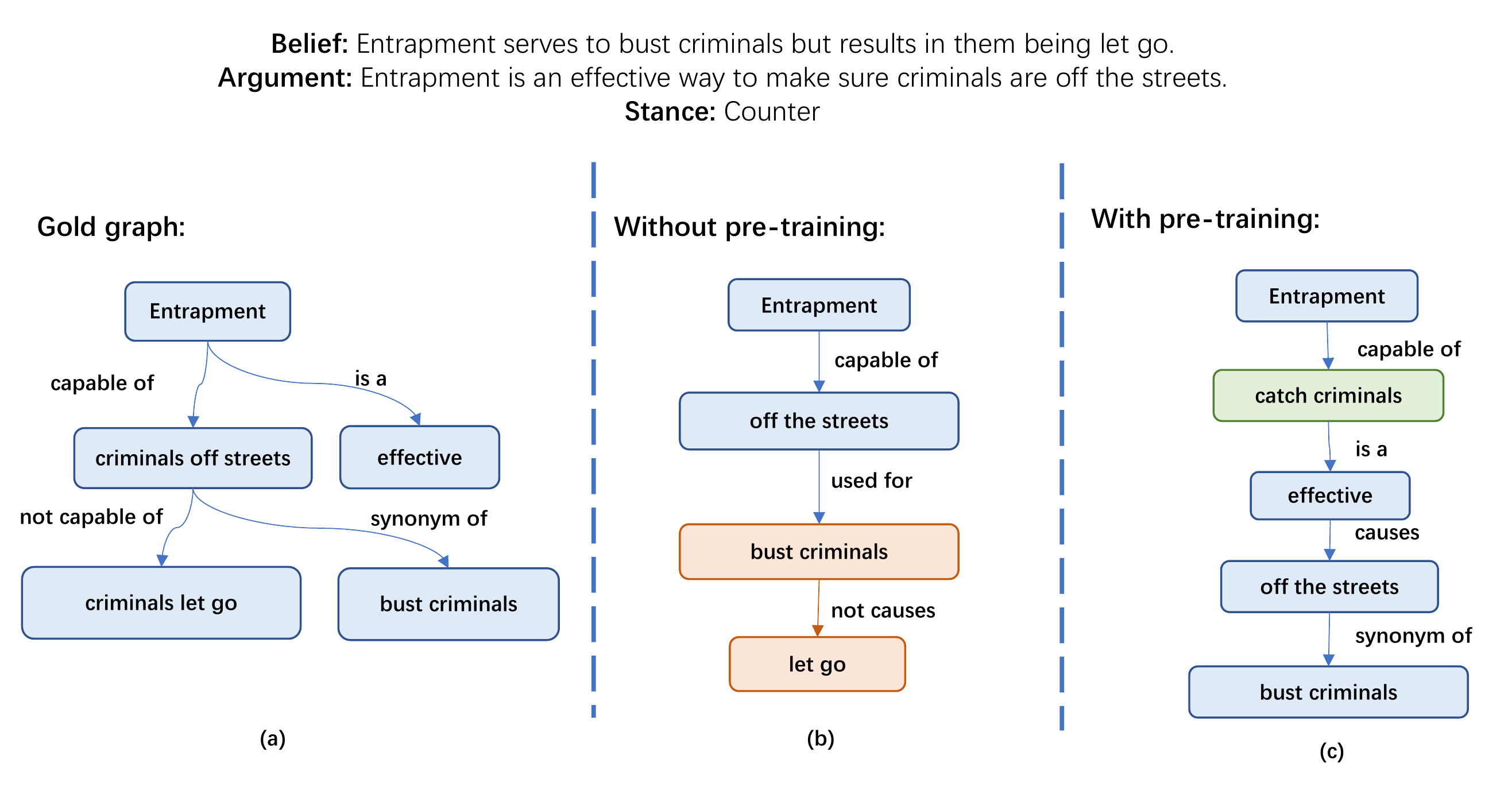}
        \caption{Example 1 on ExplaGraphs dataset.}
        \label{fig:case1}
\end{figure*}

\begin{figure*}[t]
        \centering
        \includegraphics[width=13cm]{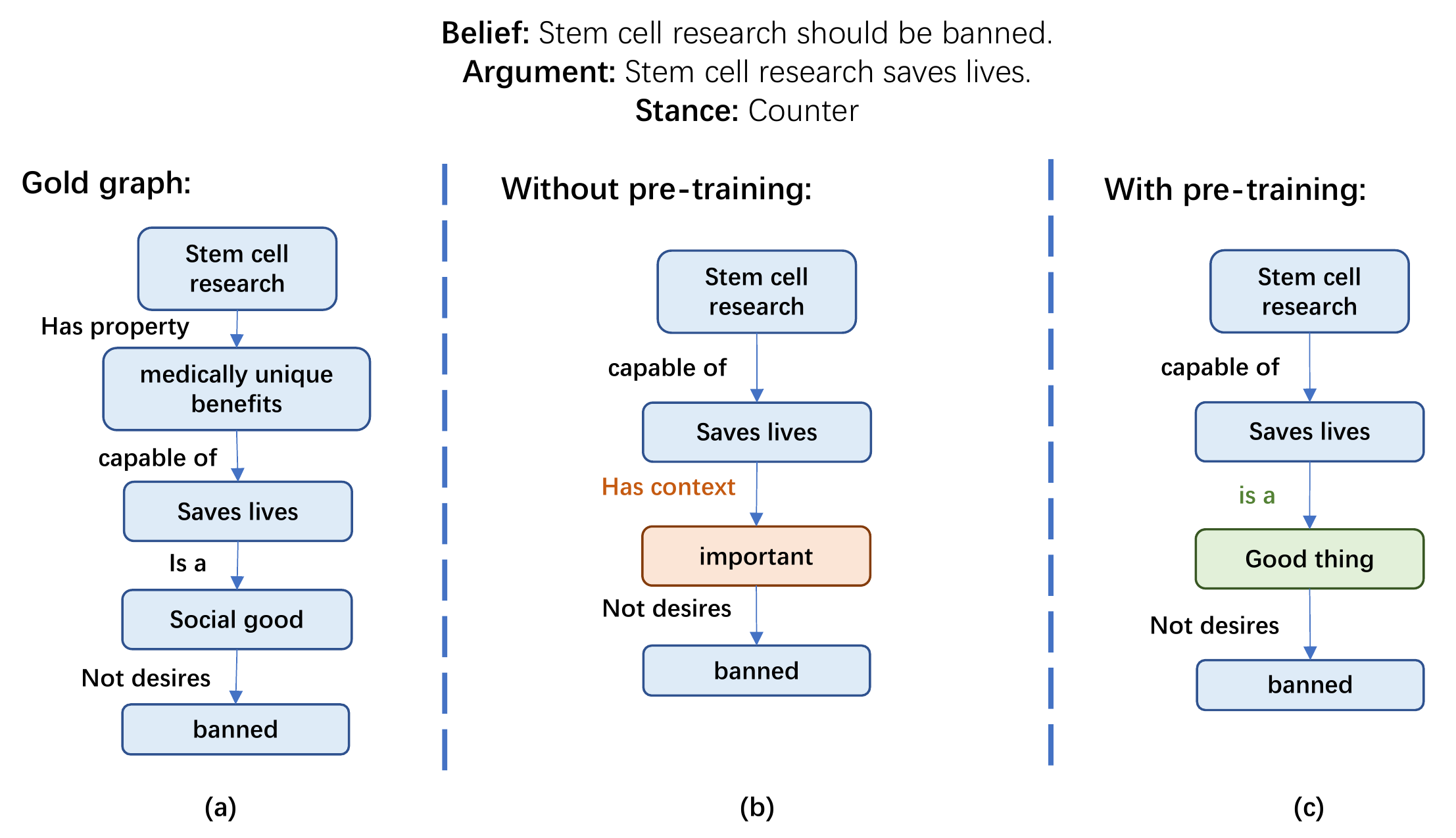}
        \caption{Example 2 on ExplaGraphs dataset.}
        \label{fig:case2}
\end{figure*}

Figure~\ref{fig:case1} and Figure~\ref{fig:case2} shows two examples on the dev set generated from our model. 
The semantic expression of the graph after pre-training is more accurate than the one without pre-training. In case 1, the graph reaches the proper concept with a longer reasoning path, and in case 2 the process of reasoning is more precise. 
Besides, comprehensively observing other examples, we found that our model is more inclined to reason step by step based on commonsense knowledge, and the reasoning process is less jumpy. 
In the golden graph, "entrapment" and "effective" are linked directly in the in the human labeling. However, in the generated graph, there is an additional node "catch criminals" between "entrapment" and "effective", refining the process. 
Moreover, in case 2 and other instances, we found the model could generate the counter relation (not desires) which is not introduced in the pre-training corpus. This indicates that our model can quickly learn new relationships from small amounts of data in downstream tasks and apply them to the graph generation process.

\subsection{Case study on CommonsenseQA and OpenbookQA}
\label{app:caseStudyCSQAOBQA}
\begin{figure*}[t]
        \centering
        \includegraphics[width=11cm]{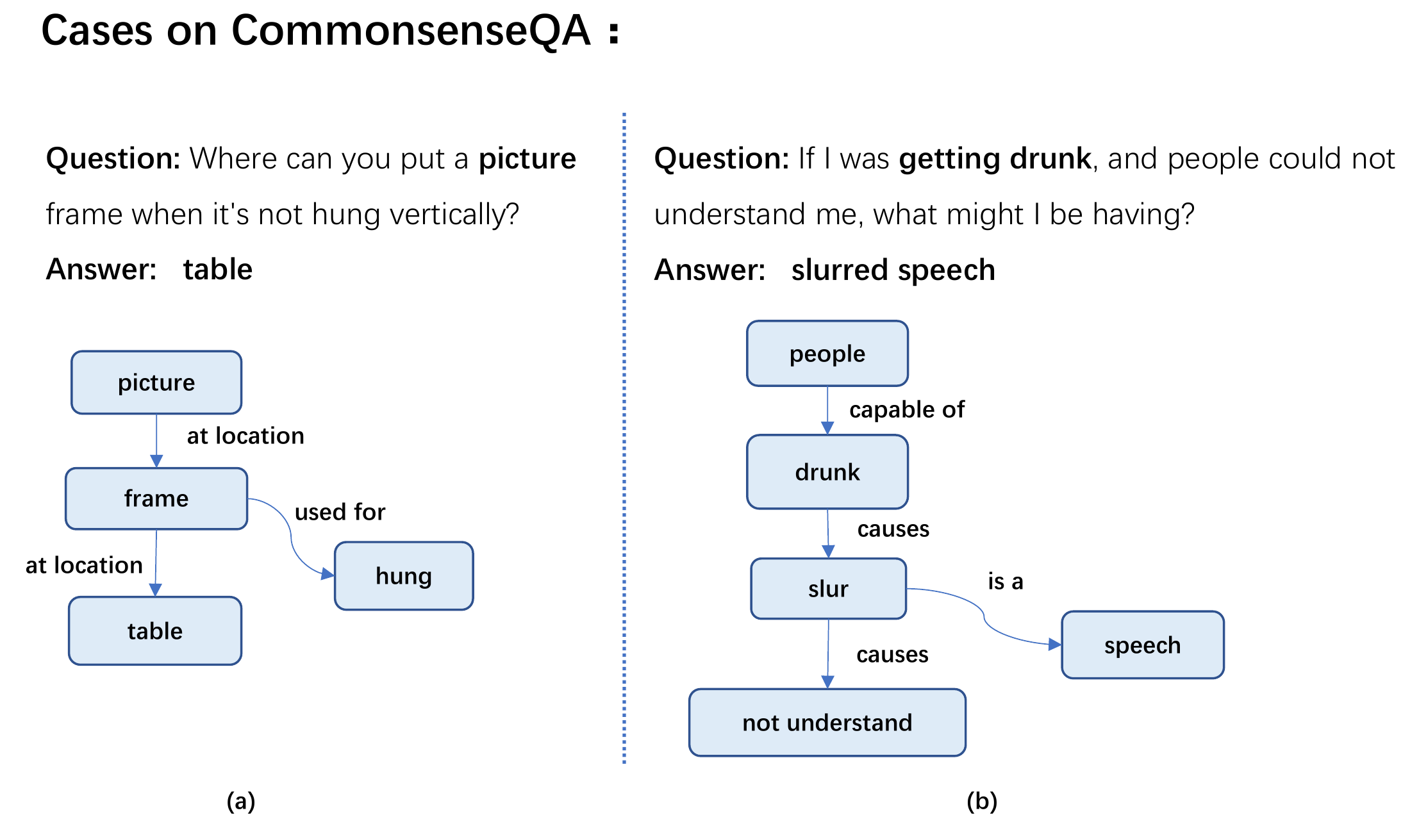}
        \caption{Graph examples generated on CommonsenseQA dataset.}
        \label{fig:csqaCase}
\end{figure*}

\begin{figure*}[t]
        \centering
        \includegraphics[width=11cm]{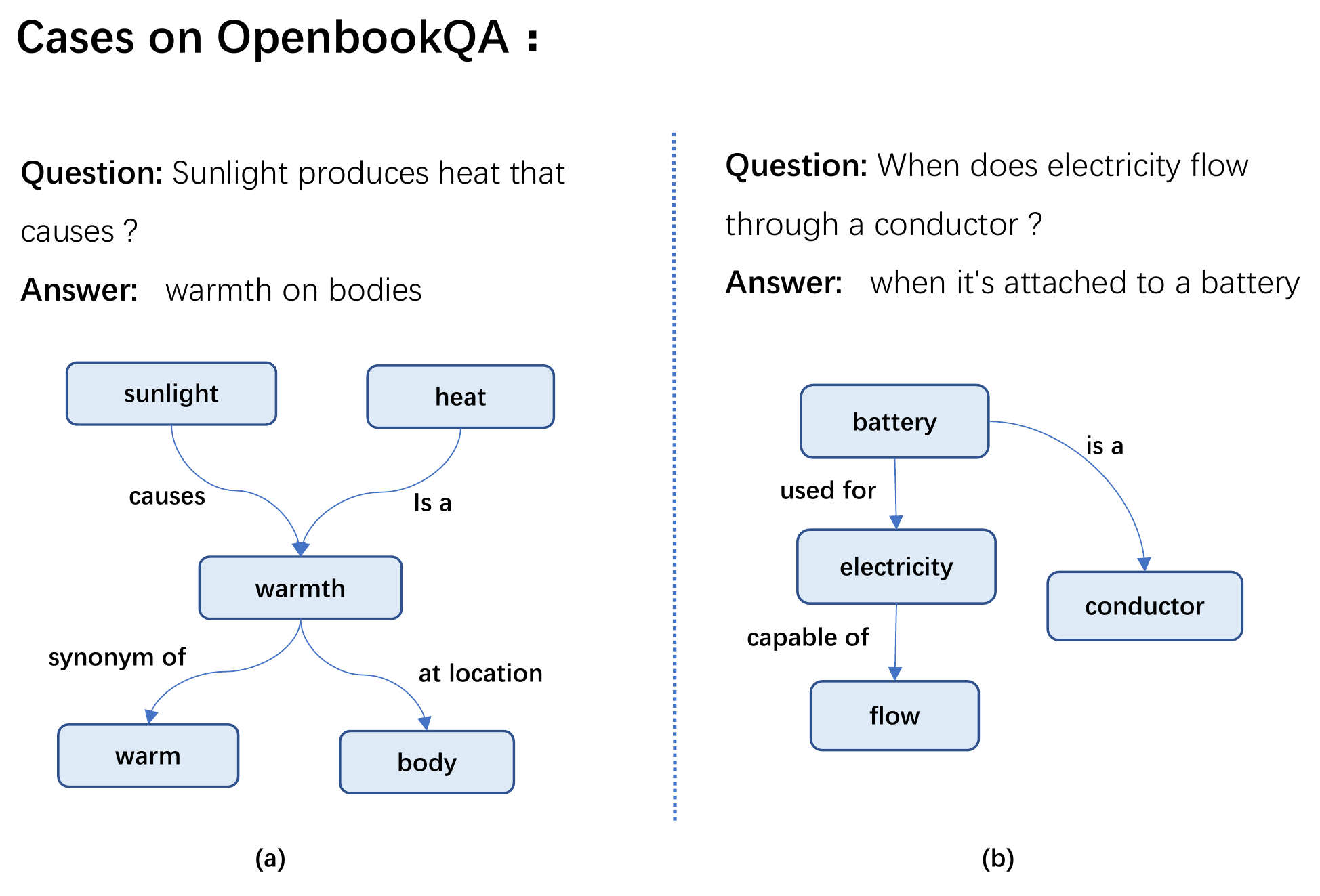}
        \caption{Graph examples generated on OpenbookQA dataset.}
        \label{fig:obqaCase}
\end{figure*}

Figure~\ref{fig:csqaCase} shows graph generated on CommonsenseQA. Figure~\ref{fig:obqaCase} shows graph generated on OpenbookQA. Observing the graphs generated, we find that our model is more inclined to generate the graph with basic commonsense knowledge rather than scientific knowledge or certain reasoning processes that are obvious to humans, especially on OpenbookQA. The model prefers to explain the reasoning process of some scientific questions with the relations contained in the pre-training corpus.

\begin{table*}[]
    \centering
    \resizebox{1.7\columnwidth}{!}{
    \begin{tabular}{m{2.5cm}<{\centering}m{6cm}m{6cm}}
\toprule 
	&	\multicolumn{1}{c}{\textbf{Input}}	&		\multicolumn{1}{c}{\textbf{Output}}		\\

\hline
\textbf{Link Prediction} & 
\texttt{predict relation: product | numbers} &
\texttt{relatedto}
\\
\hline
\textbf{Tail Prediction} & 
\texttt{predict tail: attention | causes}    &
\texttt{make\_people\_laugh}\\
\hline
\hline

\ours-\textbf{Easy}	&	\texttt{eating\_quickly | confetti [SEP] [ANSWER] is a result of eating\_quickly and confetti is used for [I\_E] then [ANSWER] is a subevent of [I\_E] ? [SEP] eating\_quickly : causes : eating\_too\_much | confetti : usedfor : celebrating | celebrating : hassubevent : eating\_too\_much | celebrating : hassubevent : fireworks | celebrating : hassubevent : eating\_too\_much}	&	\texttt{eating\_quickly : causes : eating\_too\_much | confetti : usedfor : celebrating | celebrating : hassubevent : eating\_too\_much}\\
\hline
\ours-\textbf{Normal}	&	\texttt{eating\_quickly | confetti [SEP] eating\_quickly | confetti [SEP] What eating\_quickly causes and confetti is used for then is a subevent of ? [SEP] eating\_quickly : causes : eating\_too\_much | confetti : usedfor : celebrating | celebrating : hassubevent : eating\_too\_much | eating\_quickly : usedfor : dogs | dogs : capableof : trained | carnival : isa : celebrating | pub : usedfor : celebrating | eating\_too\_much : causes : getting\_fat}	&	\texttt{eating\_quickly : causes : eating\_too\_much | confetti : usedfor : celebrating | celebrating : hassubevent : eating\_too\_much}\\
\hline
\ours-\textbf{Hard}	&	\texttt{eating\_quickly | confetti [SEP] What eating\_quickly is a result of with confetti ? [SEP] eating\_quickly : causes : eating\_too\_much | confetti : usedfor : celebrating | celebrating : hassubevent : eating\_too\_much | eating\_quickly : hassubevent : soup | celebrating : hassubevent : applause | eating\_too\_much : causes : gas | gas : isa : vapor}	&	\texttt{eating\_quickly : causes : eating\_too\_much | confetti : usedfor : celebrating | celebrating : hassubevent : eating\_too\_much}	\\
\hline
\hline
\textbf{ExplaGraphs}	&	\texttt{Belief: Compulsory voting is not a good societal implementation. [SEP] Argument: Compulsory voting would allow too many uninformed people the ability to vote. [SEP] Stance: support} 	&	\texttt{(compulsory voting; causes; inefficient vote)(inefficient vote; created by; uninformed people)(uninformed people; not used for; good societal implementation)}\\
\hline
\textbf{CommonsenseQA}	&	\texttt{getting drunk | slurred speech [SEP] After getting drunk people couldn't understand him, it was because of his what?}		&	\texttt{(people; capable of; drunk) (drunk; causes; slur)(slur; is a; speech)(slur; causes; not understand)}\\
\hline
\textbf{OpenbookQA}	&	\texttt{causes | produces | heat | warmth on bodies [SEP] Sunlight produces heat that causes?}
	&	\texttt{(escope; synonym of; a telescope)(a telescope; capable of; making)(making; used for; mailing tube)}\\

\bottomrule 
\end{tabular}}
    \caption{Input samples of all the process of generating. The samples shown in the upper part is from the task of link prediction and tail prediction. The samples shown in the middle part of the table are from the pre-training task. The samples shown in the lower part are from the generation process of the downstream dataset.}
    \label{tab:inputSample}
\end{table*}

\begin{table*}
\centering
\resizebox{\textwidth}{!}{
\begin{tabular}{lccccccccc}
\toprule 
& \textbf{3000} & \textbf{0.03M} & \textbf{0.3M} & \textbf{1M} & \textbf{2M} & \textbf{3M} & \textbf{10M} & \textbf{20M} & \textbf{Overall}\\
\midrule
\textbf{is a} & 20.07\% & 19.84\% & 19.68\% & 19.71\% & 19.73\% & 19.69\% & 19.71\% & 19.70\% & 19.70\%\\
\midrule
\textbf{at location} & 11.84\% & 12.43\% & 12.58\% & 12.53\% & 12.51\% & 12.57\% & 12.55\% & 12.55\% & 12.55\%\\
\midrule
\textbf{part of} & 4.57\% & 4.53\% & 4.69\% & 4.68\% & 4.68\% & 4.68\% & 4.67\% & 4.67\% & 4.67\%\\
\midrule
\textbf{capable of} & 6.17\% & 5.96\% & 5.92\% & 5.89\% & 5.90\% & 5.91\% & 5.91\% & 5.91\% & 5.91\%\\
\midrule
\textbf{has context} & 3.46\% & 3.85\% & 3.79\% & 3.80\% & 3.81\% & 3.80\% & 3.81\% & 3.81\% & 3.81\%\\
\midrule
\textbf{desires} & 1.26\% & 1.21\% & 1.30\% & 1.31\% & 1.31\% & 1.30\% & 1.31\% & 1.31\% & 1.31\%\\
\midrule
\textbf{antonym} & 13.28\% & 12.75\% & 12.87\% & 12.89\% & 12.90\% & 12.91\% & 12.92\% & 12.91\% & 12.91\%\\
\midrule
\textbf{used for} & 10.51\% & 10.14\% & 10.18\% & 10.18\% & 10.22\% & 10.19\% & 10.18\% & 10.19\% & 10.19\%\\
\midrule
\textbf{causes} & 9.75\% & 10.45\% & 10.28\% & 10.38\% & 10.35\% & 10.35\% & 10.35\% & 10.35\% & 10.35\%\\
\midrule
\textbf{has subevent} & 13.78\% & 13.44\% & 13.39\% & 13.35\% & 13.33\% & 13.31\% & 13.32\% & 13.31\% & 13.32\%\\
\midrule
\textbf{has property} & 2.86\% & 2.66\% & 2.62\% & 2.63\% & 2.60\% & 2.61\% & 2.61\% & 2.61\% & 2.61\%\\
\midrule
\textbf{receives action} & 0.99\% & 1.22\% & 1.15\% & 1.14\% & 1.13\% & 1.14\% & 1.13\% & 1.14\% & 1.14\%\\
\midrule
\textbf{made of} & 0.35\% & 0.34\% & 0.38\% & 0.37\% & 0.37\% & 0.38\% & 0.37\% & 0.38\% & 0.38\%\\
\midrule
\textbf{not desires} & 0.89\% & 0.95\% & 0.91\% & 0.91\% & 0.91\% & 0.91\% & 0.91\% & 0.91\% & 0.91\%\\
\midrule
\textbf{created by} & 0.12\% & 0.16\% & 0.17\% & 0.17\% & 0.16\% & 0.17\% & 0.16\% & 0.16\% & 0.16\%\\
\midrule
\textbf{not capable of} & 0.10\% & 0.08\% & 0.08\% & 0.08\% & 0.08\% & 0.08\% & 0.08\% & 0.09\% & 0.08\%\\
\midrule
\textbf{Total Triples}	&	6039	&	60437&	604311	&	2014711	&	4028880	&	6042135	&	20145322	&	40289953	&	73191788\\
\bottomrule 
\end{tabular}
}
\caption{Distribution of relations in corpora of different sizes. }
\label{tab:relationDistribution}
\end{table*}

\begin{table*}
\centering
\resizebox{.98\columnwidth}{!}{
\begin{tabular}{cc}
\toprule 
\textbf{Relation} & \textbf{Template} \\ 
\midrule
\multirow{3}{*}{[X, antonym, Y]} & X is opposite to Y\\
\multirow{3}{*}{} & Y is opposite to X\\
\multirow{3}{*}{} & X is the opposite of Y\\
\midrule
\multirow{9}{*}{[X, atlocation, Y]} & X is located in Y\\
\multirow{9}{*}{} & X , which is located in Y\\
\multirow{9}{*}{} & X, located in Y\\
\multirow{9}{*}{} & X has the position of Y\\
\multirow{9}{*}{} & X, who has the position of Y\\
\multirow{9}{*}{} & X, whose position is that of Y\\
\multirow{9}{*}{} & X's position is Y\\
\multirow{9}{*}{} & X, who holds the position of Y\\
\multirow{9}{*}{} & X holds the position of Y\\
\midrule
\multirow{5}{*}{[X, capableof, Y]} & X is capable of Y\\
\multirow{5}{*}{} & X can Y\\
\multirow{5}{*}{} & X has the ability of Y\\
\multirow{5}{*}{} & Y is the ability of X\\
\multirow{5}{*}{} & Y can be done by X\\
\midrule
\multirow{6}{*}{[X, causes, Y]} & X causes Y\\
\multirow{6}{*}{} & X is a cause of Y\\
\multirow{6}{*}{} & Y because X\\
\multirow{6}{*}{} & Y is because of X\\
\multirow{6}{*}{} & X has a result of Y\\
\multirow{6}{*}{} & Y is a result of X\\
\midrule
\multirow{4}{*}{[X, createdby, Y]} & X is created by Y\\
\multirow{4}{*}{} & Y created X\\
\multirow{4}{*}{} & X is made by Y\\
\multirow{4}{*}{} & Y made X\\
\midrule
\multirow{3}{*}{[X, isa, Y]} & X is a Y\\
\multirow{3}{*}{} & X is also a Y\\
\multirow{3}{*}{} & X is equal to Y\\
\midrule
\multirow{4}{*}{[X, desires, Y]} & X desires Y\\
\multirow{4}{*}{} & X wants Y\\
\multirow{4}{*}{} & Y is desired by X\\
\multirow{4}{*}{} & Y is wanted by X\\

\midrule
\multirow{2}{*}{[X, hassubevent, Y]} & X has a subevent of Y\\
\multirow{2}{*}{} & Y is a subevent of X\\
\midrule
\multirow{3}{*}{[X, partof, Y]} & X is part of Y\\
\multirow{3}{*}{} & X is a part of Y\\
\multirow{3}{*}{} & X, which is part of Y\\
\midrule
\multirow{4}{*}{[X, hascontext, Y]} & X has context of Y\\
\multirow{4}{*}{} & X has a context including Y\\
\multirow{4}{*}{} & when talking about X, we also talking about Y\\
\multirow{4}{*}{} & X is close to Y in context\\
\midrule
\multirow{3}{*}{[X, hasproperty, Y]} & X has a property of Y\\
\multirow{3}{*}{} & Y is a property of X\\
\multirow{3}{*}{} & X, with a property of Y\\
\midrule
\multirow{4}{*}{[X, madeof, Y]} & X is made of Y\\
\multirow{4}{*}{} & Y is used to make X\\
\multirow{4}{*}{} & X's material is Y\\
\multirow{4}{*}{} & the material of X is Y\\
\midrule
\multirow{6}{*}{[X, notcapableof, Y]} & X is not capable of Y\\
\multirow{6}{*}{} & X can not Y\\
\multirow{6}{*}{} & Y can't be done by X\\
\multirow{6}{*}{} & X doesn't has the ability of Y\\
\multirow{6}{*}{} & X is not able that Y\\
\multirow{6}{*}{} & Y is not a ability of X\\
\midrule
\multirow{5}{*}{[X, notdesires, Y]} & X doesn't desire Y\\
\multirow{5}{*}{} & X don't want X\\
\multirow{5}{*}{} & X don't desire Y\\
\multirow{5}{*}{} & X doesn't want Y\\
\multirow{5}{*}{} & X doesn't need Y\\
\midrule
\multirow{3}{*}{[X, receivesaction, Y]} & X receive an action of Y\\
\multirow{3}{*}{} & Y will give an action to X\\
\multirow{3}{*}{} & when Y, X will receive an action\\
\midrule
\multirow{2}{*}{[X, usedfor, Y]} & X is used for Y\\
\multirow{2}{*}{} & Y will use X\\ 
\bottomrule   

\end{tabular}
}
\caption{All the templates used to converting the triple to natural language expression. X means the head concept and Y means the tail concept.}
\label{tab:templates}
\end{table*}


\end{document}

%% file: sections/introduction.tex
\section{Introduction}

\begin{figure}[t]
        \centering
        \includegraphics[width=7.5cm]{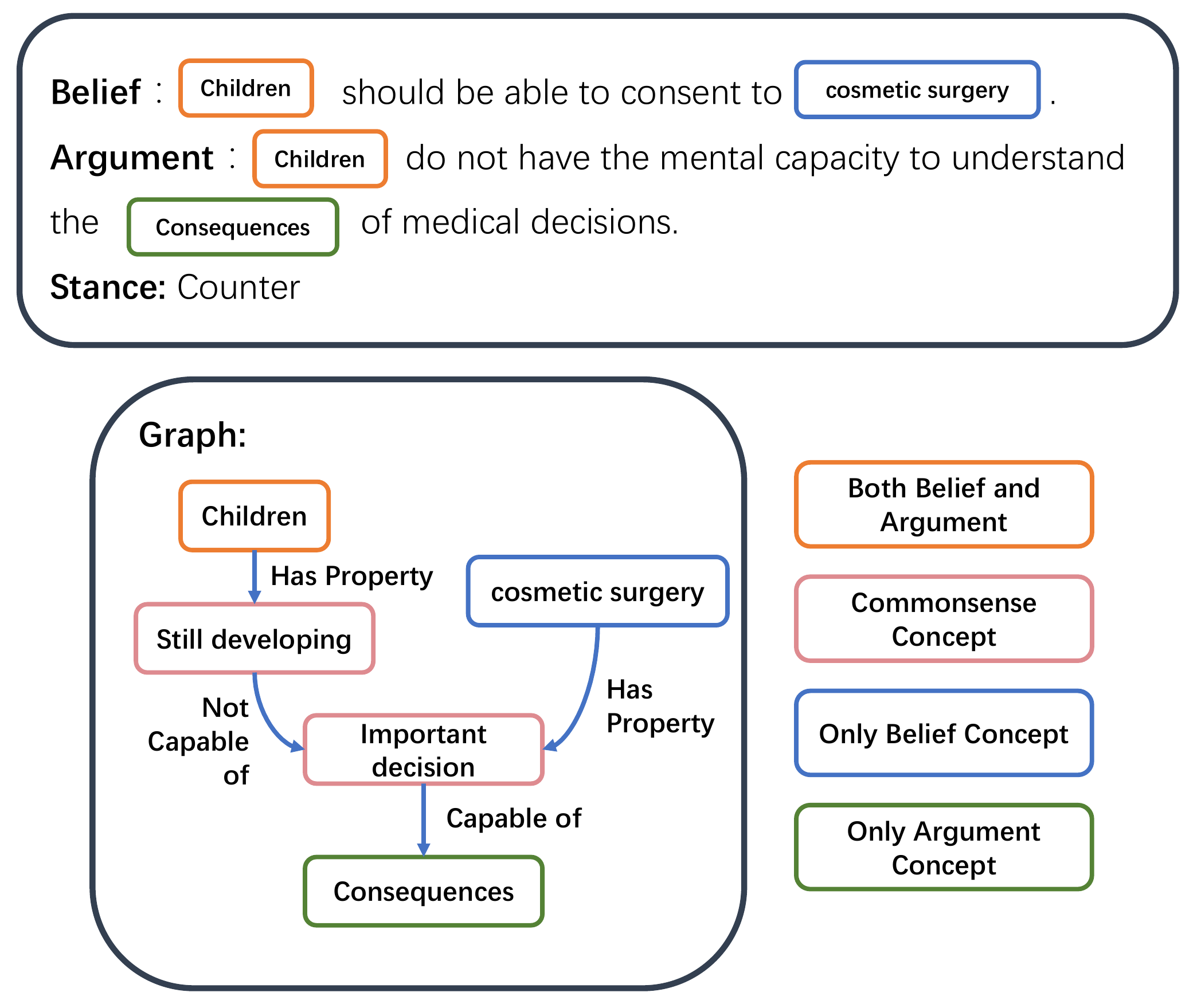}
        \caption{An example of the task of explanation graph generation (from ExplaGraphs dataset). Given a piece of natural text, the model needs to generate a graph depicting the reasoning process. }
        \label{fig:example}
\end{figure}

Generating an explanation to probe why the model obtains answers is a long-term goal in the development of intelligent systems, especially in reasoning-related tasks, such as E-SNLI\cite{camburu2018snli}, ECQA\cite{aggarwal-etal-2021-ecqa}, HotpotQA\cite{yang-etal-2018-hotpotqa} and ExplaGraphs\cite{saha-etal-2021-explagraphs}. 
According to the types of explanations, existing explanation generation tasks can be mainly divided into three types, including textual highlights explanation generation \cite{yang-etal-2018-hotpotqa, camburu2018snli}, natural language explanation generation \cite{camburu2018snli, https://doi.org/10.48550/arxiv.2010.12762, inoue-etal-2021-summarize} and structured explanation generation \cite{xie-etal-2020-worldtree, saha-etal-2021-explagraphs}.
Among all these tasks, structured explanation generation achieve growing attention recently since the explanation in this task is usually a graph, which is clean enough, and easy to evaluate from the perspective of structure and semantics (denoted as an explanation graph). 
An example of a structured explanation generation task is shown in Figure~\ref{fig:example}.

Pre-trained language models, such as RoBERTa \cite{roberta}, BART \cite{lewis-etal-2020-bart} and T5 \cite{t5} have demonstrated their powerful capabilities in a great many language understanding tasks.
As a result, when it comes to explanation graph generation, existing studies primarily fine-tune pre-trained language models on downstream tasks directly \cite{xie-etal-2020-worldtree, saha-etal-2020-prover, saha-etal-2022-explanation}. 
While typical pre-trained language models (PLMs) are pre-trained on textual corpus only, fine-tuning on downstream tasks directly may lead to a significant discrepancy between text-based language models and explanation graphs. 
To mitigate this issue, we argue that pre-training over task data can be an ideal way to bridge the above gap. Such a pre-training manner can be a subtle solution to inject inductive bias into PLMs. 
However, the scale of existing datasets is relatively small since it costs a lot to label explanation graphs, and pre-training on existing data is insufficient to bridge the gap. 
To this end, an appealing solution is to continually pre-train PLMs on a large scale automatically synthetic corpus containing explanation graphs instead of human labeling before fine-tuning. 
The explanation graph is highly structured and contains diverse entities and relations, which is easily synthesized by randomly assigning different values to each entity and relation positions.

\begin{figure}[t]
        \centering
        \includegraphics[width=7.5cm]{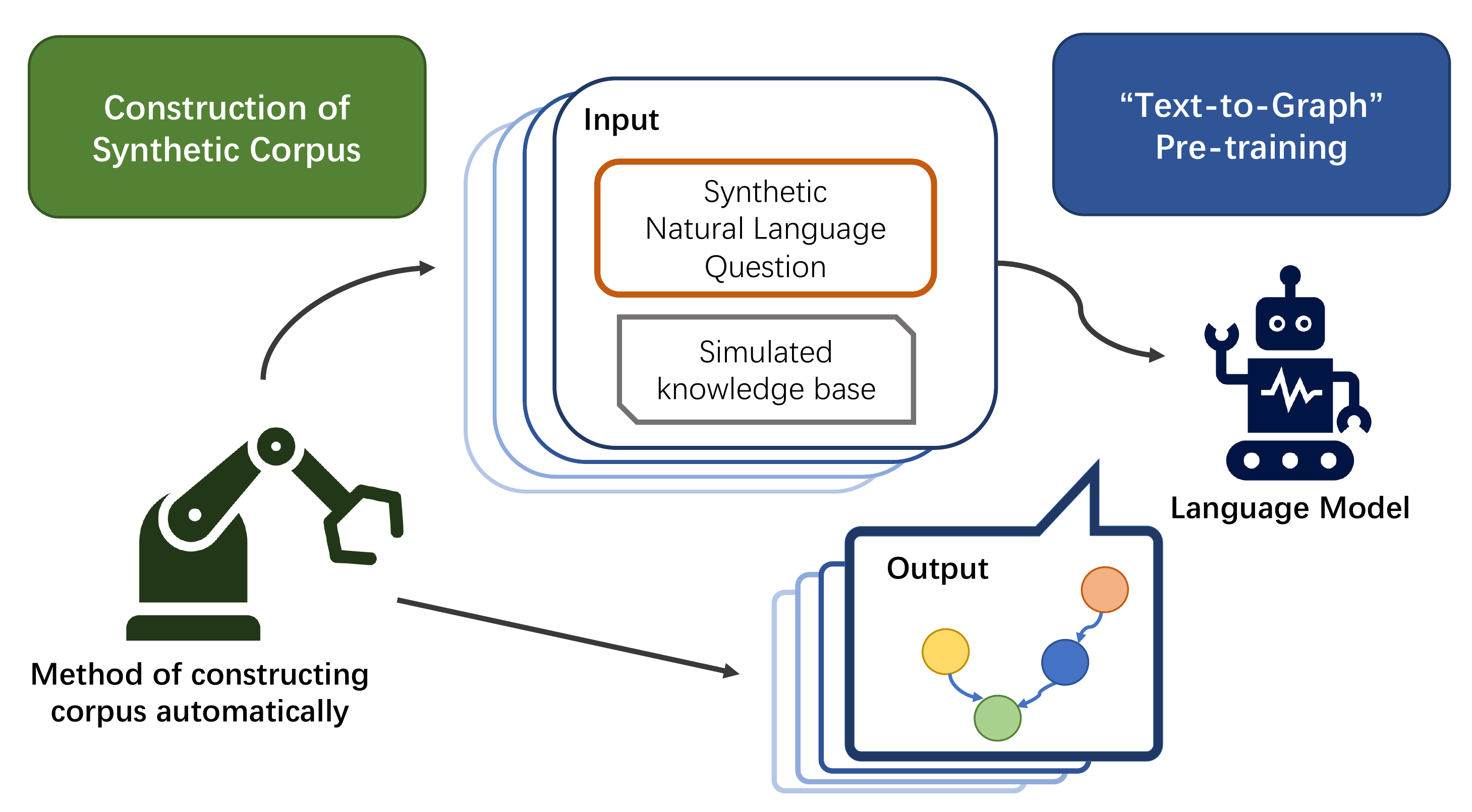}
        \caption{The overview of \textsc{\ours}. The model is first pre-trained on a large amount of synthetic data in the form of "text2graph", and then fine-tuned on a downstream task with a small amount of data.}
        \label{fig:overview}
\end{figure}


In this paper, we propose \ours\ (\textbf{E}xplanation \textbf{G}raph \textbf{G}eneration via \textbf{G}enerative \textbf{P}re-training over Synthetic Graphs), a novel pre-trained framework for explanation graph generation. 
Specifically, as shown in Figure~\ref{fig:overview}, \ours\ is composed of two key components: the "Text-to-Graph" pre-training task and the construction of the pseudo training data. Different from previous natural language-based pre-training tasks, the “Text-to-Graph” task takes external knowledge sources and questions containing partial reasoning progress information as input, and its target is to generate relevant explanation graphs. In addition, to avoid the high cost of retrieving graphs for the simulated questions from the knowledge base, we propose a novel approach to constructing questions from simulated graphs, which automatically constructs a large amount of pseudo data.

Experiment results on the ExplaGraphs benchmark demonstrate that our approach could improve the ability of the model to generate explanatory graphs significantly.
Moreover, the model also shows excellent graph generation ability on other reasoning datasets.

Overall, we make the following key contributions:

\begin{itemize}
    \item We propose a novel pre-training task by mapping the input question to a structural explanation graph, which guides the model to learn the connections between natural language questions and structured graphs.
    \item We propose a novel approach to synthesize corpus by automatically constructing structured graphs and queries to form the large-scale corpus.
    \item Among the models with similar scales, our model achieves competitive results. Furthermore, the results of our experiments indicate that our model is capable of producing acceptable graphs on reasoning datasets without labeled graphs.
\end{itemize}

%% file: sections/Background_overview.tex
\section{Overview and Background}

In this paper, we concentrate on the task of explanation graph generation. An example is depicted in Figure~\ref{fig:example}. Given a piece of natural language text $T$, the model needs to generate a graph $G$ which encapsulates the internal reasoning path of the input text. The specific content of the input $T$ 
 is contingent upon the specific downstream tasks (belief + augment + stance in stance prediction, question + answer in QA, etc.). For output $G$, we organize the graph into a sequence of triples in the depth-first search order.  In practice, we employ a generative model and treat graph generation as a standard text generation task.

A crucial point in this task is addressing the significant discrepancy in semantic expression structure between natural language texts and explanation inference graphs. 
An ideal way is to let the model learn this expression transfer on a large amount of natural language text and graph alignment data. However, due to the small size of labeled datasets, training on these datasets is difficult to address the issue.
Based on all of the above, we propose the  two modules of our model: the "text-to-graph" pretraining strategy introduced in section~\ref{sec:pre-training}, and the method of automatically constructing synthetic corpus introduced in section~\ref{sec:corpus}.

%% file: sections/2_The_Text2graph_Pre-training_Strategy.tex
\section{The Text2graph Pre-training Strategy}
\label{sec:pre-training}
\begin{figure}[t]
        \centering
        \includegraphics[width=7cm]{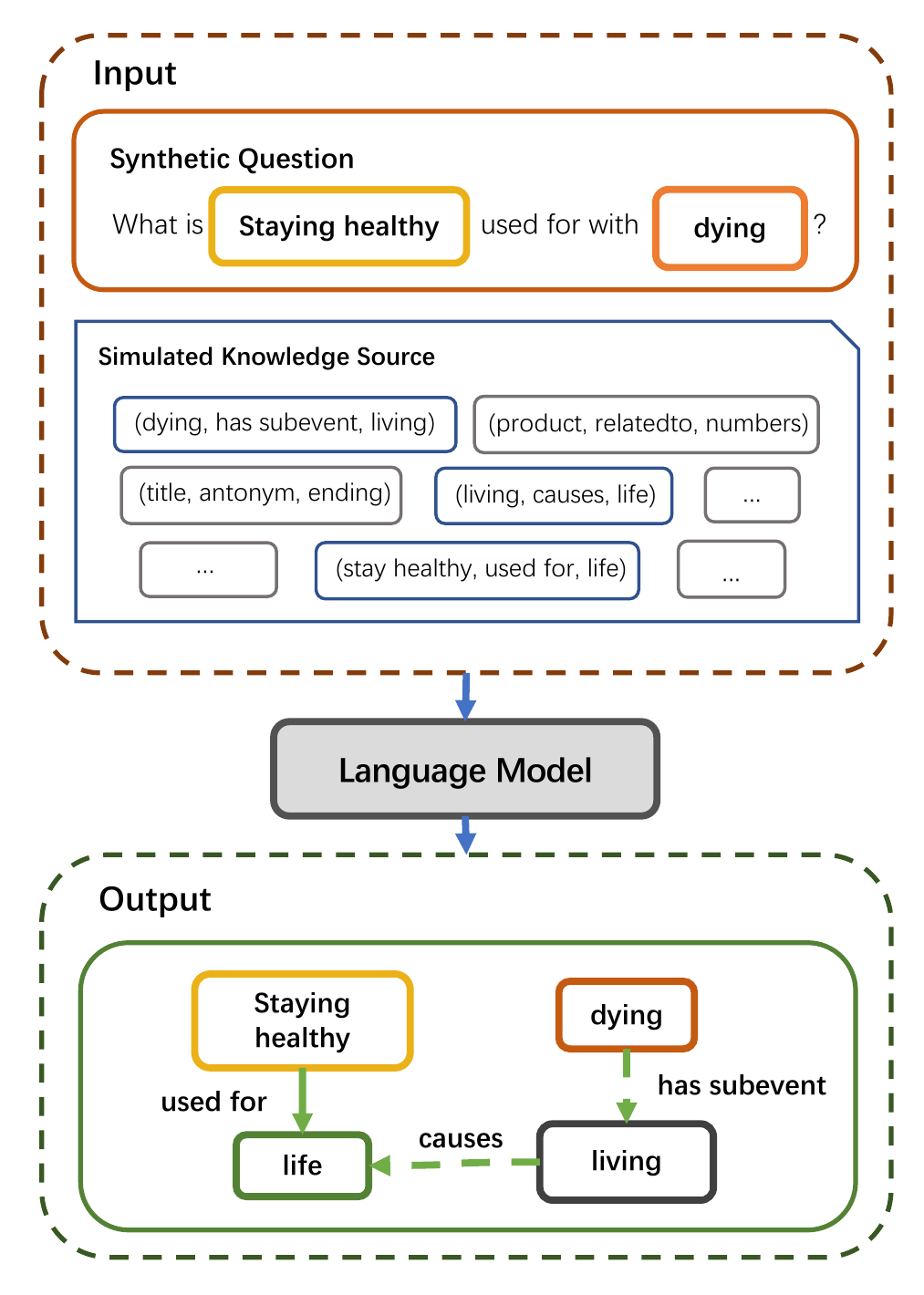}
        \caption{The illustration of "text-to-graph" task. The input comprises the synthetic question and a simulated knowledge source. In the simulated knowledge source, the triple related to the reasoning is marked as blue and the one not related is in grey. In the training process, the triples in the knowledge source will be randomly shuffled and will not be marked as relevant or not. }
        \label{fig:pretraining}
\end{figure}
Typical pre-training strategies of various PLMs are based on the corpus of natural language text (e.g. MLM and NSP(BERT), text denoising(BART), and text-to-text generation(T5)). However, in the explanation graph generation task, the explanation graph is different from the natural language text both in the representation form and in the semantic structure, which leads to a huge gap between the two kinds of representations. Apart from typical tasks, some pre-training strategies are applied to recover the triples in the knowledge graph for knowledge graph question answering(KGQA)\cite{saxena-etal-2022-kgt5}. However, such a pre-training method is not able to cover the explanation graph generation task due to the separation of pre-training steps between structured corpus and natural language, which is unable to bridge the gap between natural language and structured graphs. 

Since the explanation graph generation task is required to translate natural language text into a graph, we believe the key point is to map the natural language texts to structured graphs in the learning process implicitly. To this end, we set the form of the pre-training task as "text-to-graph" in \ours. 
As depicted in Figure~\ref{fig:pretraining}, the format of the pre-training task is analogous to that of a normal generation task. 
Given a query in natural language question and a simulated knowledge source, the model concatenates the two part together as input and generate a sequence of triples representing the reasoning graph from the query to the answer. 
By learning aligned "text-to-graph" pairs, the model acquires text-to-graph mapping in the process, and its capability for structured text generation is also enhanced. 
Real input samples are presented in Appendix~\ref{app:inputexampleappendix} for further reference.

The query and the graph of the answer come from the auto-construction method we propose, which will be discussed in the next section. 
To construct the simulated knowledge source (a collection of triples), we take the triples of the gold answer as a starting point and add random triples that are not relevant to the reasoning process to disrupt the collection. The final size of the simulated knowledge source is approximately 1.5 to 2 times the length of the graph.

%% file: sections/3_The_Construction_of_Synthetic_Corpus.tex
\section{The Construction of Synthetic Corpus}
\label{sec:corpus}
Pre-training tasks necessitate the support of large-scale corpus.
However, all the existing datasets with human-labeled graphs are small in scale due to the high cost of manually annotating, which is not enough to support the pre-training process. 
\begin{figure*}[t]
        \centering
        \includegraphics[width=16cm]{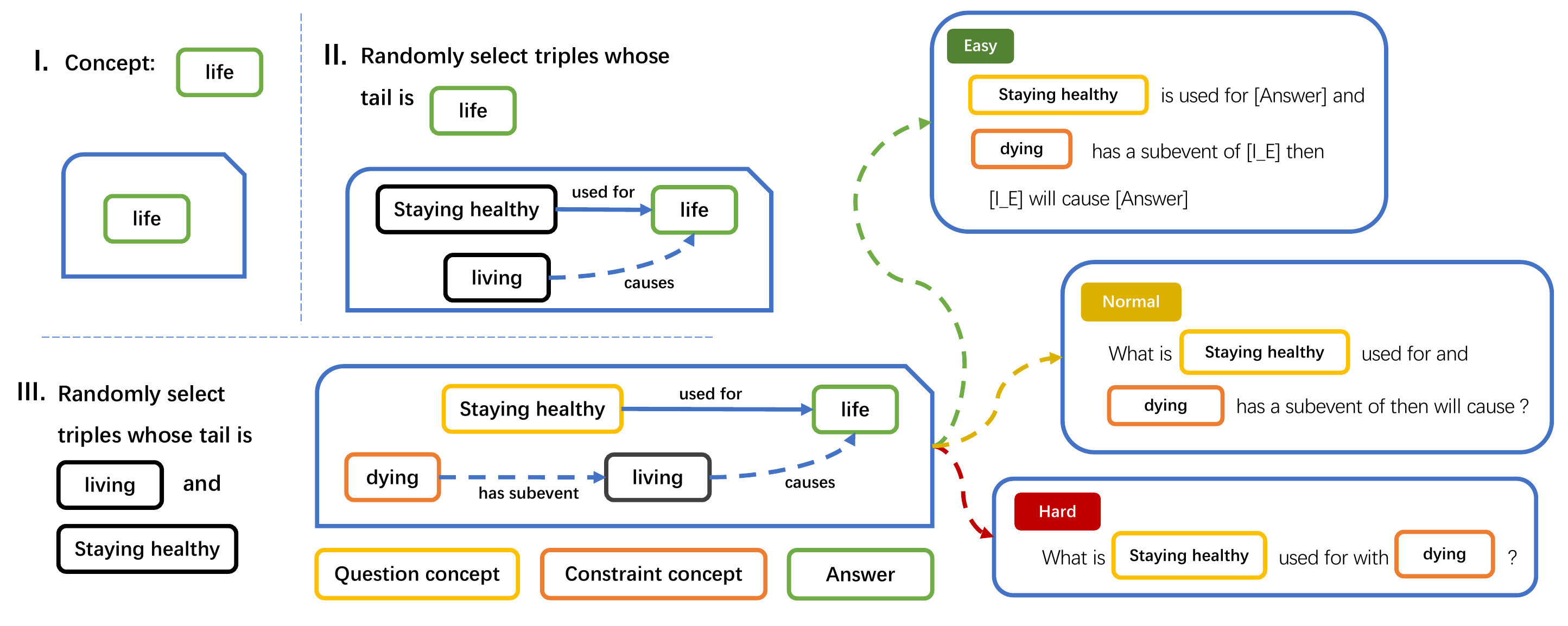}
        \caption{The construction of graphs and queries. The left part introduces the construction of the graphs. Given a node, we retrieve triples step by step forward, and then randomly select several triples (the number is allowed to be 0, 1, or 2) for concatenation. The right part describes the construction of the queries. For each graph, we construct three different queries with different difficulties, containing different amounts of information and in different forms.}
        \label{fig:corpusConstruction}
\end{figure*}
To address this issue, we propose an automatic method of constructing the pair of the natural language query and the explanation reasoning graph. 
The conventional way to get a graph from a piece of natural language text is to search in the external knowledge base. However, the complexity of searching would increase exponentially with the number of nodes and the length of edges in graphs. 
Therefore, we invert this process, synthesizing a reasoning graph first and then constructing a query based on the graph next.

\subsection{The Synthesizing of the Graph}
Observing the reasoning process of the downstream reasoning tasks, it is evident that the reasoning path of a specific instance not solely depends on the problem entity as the starting point of reasoning, but also depends on other entities in the problem as constraints, which ultimately lead to the sink point. So we construct the explanation graph from back to front to ensure that there is only one sink (i.e. the answer) in the graph and the relationship of each edge is used.

The process of construction is shown in Figure \ref{fig:corpusConstruction}. Initially, a concept is randomly selected as the sink of the graph (also the answer to the query in the following steps). Subsequently, triples are retrieved recursively, and a random number of them (ranging from 0 to 2) are incorporated into the graph. All the triples are retrieved from ConceptNet\cite{speer2017conceptnet}, which is an external knowledge base containing concepts as nodes and relations as edges.
Additionally, the relationship “relatedTo” is prevalent among the concepts, which will seriously affect the reasoning process, so it is deleted. Furthermore, certain other relations are merged, resulting in a total of 16 distinct relations. The distribution of the relations is introduced in Appendix~\ref{app:relationDistribution}.

\subsection{The Construction of the queries}

Inspired by the work of \citet{liu2021tapex}, we construct three queries with different difficulty levels: easy, normal, and hard for each instance of the graph, as shown in Figure~\ref{fig:corpusConstruction}.
The easy level involves retaining the start node and relation in the intermediate stages of reasoning, while hiding the sink node (which is treated as the answer) and the nodes present in the intermediate stages. The relation is then replaced with natural language annotations based on a predefined template, and the resulting triples are subsequently concatenated in their original order.
For the normal level, a similar amount of information is retained as in the easy level, but the concatenated query is further converted into a natural language expression using a predefined template, in order to simulate a realistic question-answering scenario.
For the hard difficulty level, only the start node and the first relation are retained, with all other auxiliary information removed, and the question is formulated in natural language.
All the template is shown in Appendix~\ref{app:inputexampleappendix}.

%% file: sections/4_experiments.tex
\section{Experiments}

\subsection{Datasets and metrics}

\paragraph{ExplaGraphs}
The main dataset we use is ExplaGraphs\cite{saha-etal-2021-explagraphs}, a commonsense explanation graph generation task. 
The task requires an agent to predict the stance of a given belief and argument, and subsequently generate an explanation graph that illustrates the reasoning process behind the predicted stance. 
Specifically, the explanation graph is a DAG, demonstrating the internal reasoning process between the belief and the argument. 
As shown in Figure~\ref{fig:overview},  the nodes in the graph correspond to concepts from the given text or external commonsense phrases, while the edges represent commonsense relations present in the dataset.
Each edge in the graph is an expression of one of the reasoning steps, and the ordered links of all edges provide an overall representation of the reasoning process for the user. 

In terms of the metrics, the dataset defines 6 test metrics, two of which are selected as main metrics by the prior works\cite{saha-etal-2022-explanation}: Structural Correctness Accuracy (StCA) evaluating if the graphs satisfy all structural constraints, and Semantic Correctness Accuracy (SeCA) evaluating if the graphs are both structurally and semantically correct. 
The structural constraints contain several parts: the graph should be a connected DAG, the relations belong to the relation list defined by the dataset and there are at least two concepts from the belief and two from the argument. 
The semantic correctness is evaluated by a model-based metric \cite{saha-etal-2021-explagraphs}, checking whether the semantics of the graph and the standard answer are matched. All the metrics in detail could be found in the Appendix~\ref{app:metrics}.


\paragraph{Other reasoning datasets}

To prove the generalization ability of the model, we also conducted experiments on two other general commonsense reasoning datasets in addition to ExplaGraphs: CommonsenseQA\cite{talmor-etal-2019-commonsenseqa} and OpenbookQA\cite{mihaylov-etal-2018-openbookqa}. 
CommonsenseQA is a 5-way multiple-choice question answering dataset that focuses on commonsense reasoning, while OpenBookQA is a 4-way multiple-choice question answering dataset that requires reasoning with elementary science knowledge.
Since there is no labeled commonsense reasoning graph on these datasets, we evaluate the results of the dev set of these two datasets manually from the point of semantics and analyze the model for specific examples. The evaluation of semantics is to check whether the semantics of the graph matches the reasoning process properly.

\subsection{Generative Baseline} \label{sec:baseline}
In line with the previous work\cite{saha-etal-2021-explagraphs, saha-etal-2022-explanation}, we generate the explanation graphs in a post-hoc manner, with a condition of the belief, the argument, and the predicted stance. In order to objectively compare the results of graph generation, the part of stance prediction in all our experiments is finished by an identical RoBERTa-based model. The first baseline model is BART, our backbone of \ours. Furthermore, we also implement other pre-training methods that have been introduced in recent studies\cite{saxena-etal-2022-kgt5} on knowledge graph question answering (KGQA), such as link prediction and tail prediction.  Link prediction is a common task in knowledge graph embedding(KGE) learning. Given two parts of a knowledge triple   (head+relation, head+tail, or relation+tail), the model is required to complete the missing element of the input. For the tail prediction task, the training process is basically the same as link prediction, but the model only needs to predict the tail entity in all instances, which is more similar to the process of step-by-step reasoning from front to back. In order to facilitate the model's understanding of the task, we add a prompt before the input triple: “Predict the head/relation/tail: xxx”. The input sample of the two tasks is shown in Appendix~\ref{app:inputexampleappendix}.

\subsection{Fine-tuning on Downstream Datasets}
For the fine-tuning process on ExplaGraphs, we follow the pipeline outlined in previous work as described above. For the fine-tuning process on CommonsenseQA and OpenbookQA, we did not use the model to generate the graph in zero-shot style, because we found that BART-Large without any learning process can hardly generate an acceptable graph in the comparison tests. To improve comparability, we fine-tune the model with the ExplaGraphs dataset before generating explanation graphs on other datasets in different groups of experiments. All the input samples are shown in Appendix~\ref{app:inputexampleappendix}.

\subsection{Experimental Setup}
The experiments include three parts: the construction of the corpus, the process of pre-training, and the process of fine-tuning. 

For corpus construction, we first synthesize 20 million reasoning graph instances and construct three questions of varying difficulty for each instance. Then, the “query-graph” pairs in three difficulty levels are mixed in equal proportion, ensuring that the total amount of data meets the experimental requirements. Except for experiments discussing the effect of the corpus scale, the scale of the corpus in other experiments is set to 0.3 million. 

For the pre-training process, we utilize the BART-Large\cite{lewis-etal-2020-bart} model in fairseq\cite{ott-etal-2019-fairseq}, a widely-employed seq2seq model that follows the standard transformer structure, as the backbone of our model. The pre-training process runs up to 50000 steps with a learning rate of 3e-5, a dropout rate of 10\%, and a max length of 1536. 

\newcommand{\dev}[1]{$^{\diamondsuit}$}
\newcommand{\test}[1]{$^{\star}$}
\begin{table*}[th]
    \centering
    \begin{tabular}{lcccccc}
    \toprule
         & SA$\uparrow$ & StCA$\uparrow$    & SeCA$\uparrow$    & G-BS$\uparrow$    & GED$\downarrow$          & EA$\uparrow$   \\
    \hline
    BART-Base\cite{saha-etal-2021-explagraphs}\dev     
                                    & 86.2  & 21.6  & 11.0  & 16.1  & 0.85  & 10.3 \\
    BART-Large\dev                  &\textbf{88.19}  &36.43  &26.13  &28.42  &0.74   &20.77  \\
    \qquad + Link Prediction\dev    &\textbf{88.19}  &40.45  &31.82  &28.39  &0.71   &14.63  \\
    \qquad + Tail Prediction\dev    &\textbf{88.19}  &41.21  &32.04  &29.15  &0.71   &22.54  \\
    \qquad + \ours\dev              &\textbf{88.19}  &\textbf{48.99}  &\textbf{37.43}  &\textbf{38.73}  &\textbf{0.65}   &\textbf{25.03}  \\
    \hline
    BART-Large\cite{saha-etal-2021-explagraphs}\test             
                                & 87.2  &34.20  &22.20	&28.90	&0.75	&20.00 \\
    Contrastive Learning \citep{saha-etal-2022-explanation}\test 
                                &87.2	&40.7	&26.30	&31.30	&0.71	&22.30 \\
    \textsc{CoCoGen}\citep{madaan2022llm}\test  
                                &87.2   &45.20  &23.74  &34.68  &0.69   &23.58 \\
    \ours\test       &\textbf{87.75}	&\textbf{50.75}	&\textbf{31.25}	&\textbf{43.86}	&\textbf{0.62}	&\textbf{27.75} \\
    \toprule
    \end{tabular}
    \caption{All the experimental results on the ExplaGraphs dataset. The line with \dev{}  is the result on the dev set. The line with \test{} is the result on the test set. For the detailed disclosure of all evaluation metrics, please refer to the Appendix~\ref{app:metrics}.}
    \label{tab:ExplaGraphsResults}
\end{table*}

For the process of fine-tuning, we build the classification model based on RoBERTa-Large\cite{roberta}, with a batch size of 32, an initial learning rate of 1e-5 with linear decay, a weight decay of 0.1, and a maximum input length of 128. The model is trained for 10 epochs. Then the fine-tuning step on ExplaGraphs for graph generation runs up to 10000 steps with a batch size of 8 and a max length of input and output of 150, keeping other parameters the same as the pre-training process. The whole training process is conducted on Nvidia-A100-40G.

%% file: sections/5_results_and_analysis.tex
\section{Results and Analysis}

\subsection{Results on ExplaGraphs}
In this section, we compare the result of our \ours\ with other baselines introduced in Sec~\ref{sec:baseline} and some released works on the same task. Following prior work~\cite{saha-etal-2022-explanation}, we report all the metrics on the ExplaGraphs dataset.

\paragraph{Effect of “Text-to-Graph” pre-training method}
\begin{table*}[th]
    \centering
    \begin{tabular}{lccccc}
    \toprule
        & StCA$\uparrow$    & SeCA$\uparrow$    & G-BS$\uparrow$    & GED$\downarrow$          & EA$\uparrow$   \\
    \hline
    BART-Large       & 36.43  & 26.13  & 28.42  & 73.84  & 20.77\\
    \hline
    \qquad + Easy    & 47.99   & 33.16  & 38.71  & 66.23  & 14.23 \\
    \qquad + Normal  & 49.5  & 33.66  & 39.56  & 64.85 & 25.1  \\
    \qquad + Hard    & 45.98  & 27.63  & 36.52  & 67.74  & 23.07  \\
    \hline
    \qquad + Mixed   & 48.99  & 37.43  & 38.73  & 65.14  & 25.03  \\
    \toprule
    \end{tabular}
    \caption{The results of the model pre-trained on a different scale of the corpus. All the results are on the dev set. As described above, we use the same classifier model in all the experiments, reaching 88.19 on SA. }
    \label{tab:difficulty}
\end{table*}

In this part, we report all the evaluation results on the dev set. As depicted in Table~\ref{tab:ExplaGraphsResults}, our pre-training method in \ours\ improves StCA by 12.56\% and SeCA by 11.3\% compared to BART-Large without “text-to-graph” pre-training, indicating our method could significantly enhance the model's capability for graph generation in terms of both structure and semantic understanding. 

Furthermore, based on the same backbone model, the pre-training method in \ours\ also outperforms other listed pre-training methods in the table across all the metrics, as evident in Table~\ref{tab:ExplaGraphsResults}, which demonstrates the efficacy of our modeling approach. The gains on the task of link prediction and tail prediction are not relatively significant on structural correctness and semantic correctness, which means the aligned input pair of “text-graph” and the output of graph is crucial for the model to learn the mapping between natural language text and structural graph. The case study is discussed in Appendix~\ref{app:caseStudyExplaGraphs}.

\paragraph{Comparison with other works}
In this part, we compare our results with some other representative results on the ExplaGraphs dataset. 
\begin{itemize}
    \item \citet{saha-etal-2022-explanation} proposes some methods to construct structurally and semantically positive and negative graphs and leverages these graphs in different contrastive learning models. In order to make a fair comparison, we take the results of this method on BART-Large. 
    \item \textsc{CoCoGen}\cite{madaan2022llm} treats the structured commonsense reasoning task as a code generation task and uses a code generation language model CODEX\cite{https://doi.org/10.48550/arxiv.2107.03374} to generate the graph with few-shot prompting. There are also other results of the same method on different natural language large language models(LLMs), such as CURIE and DAVINCI. We only compare with the best result of them.
\end{itemize}

The results of the test set are summarized in Table~\ref{tab:ExplaGraphsResults}. The comparison demonstrates that our proposed method, \ours, outperforms both of the aforementioned methods, particularly in terms of semantic correctness accuracy (SeCA).  The results show that the pre-training method on aligned “text-graph” pair could help the model learn the mapping between natural language and graphs better than training on a single downstream task. Besides, specific pre-training methods could also endow small models with a better ability of semantic understanding on the specific task (graph generation here) than large language models.

\subsection{Other Analysis}

\paragraph{Effect of the difficulty of the query}
In \ours\, we construct a query in three different difficulties and mix the corpus in the main experiment as multi-task training. Table~\ref{tab:difficulty} shows the results on different queries. It is significant  that the utilization of a mixed corpus leads to a more substantial improvement than training on a single sub-task alone.
Due to the same graph generation form, the structural accuracy(StCA) of all sub-task is improved significantly; the benefits brought by the mixed corpus are mainly reflected in the semantic accuracy(SeCA).

A comparison of different sub-tasks reveals that the results for queries of normal difficulty are the most favorable. 
The queries in normal difficulty retain the form of a natural language compared to easy and retain more intermediate reasoning information compared to hard. 
This suggests that, in the training process based on a large-scale synthetic corpus, the closer the training task is to the downstream task and the simpler it is, the better the model learns.

The model pre-trained on simple corpus demonstrates superior performance in comparison to the one based on the easy corpus. Compared to easy difficulty, the pair of simple query and graph has a form that is more congruent to the explanation graph generation task. This finding aligns with previous work\cite{devlin-etal-2019-bert}, which suggests that pre-training on a task that is more closely aligned to the downstream task leads to improved performance. Besides, the model pre-trained on simple corpus also outperforms the one based on the hard corpus, despite the fact that both present the same form. This highlights the importance of selecting an appropriate difficulty level for pre-training tasks in order to achieve optimal efficiency.

\paragraph{Effect of the scale of corpus}
\begin{figure}[h]
        \centering
        \includegraphics[width=7.5cm]{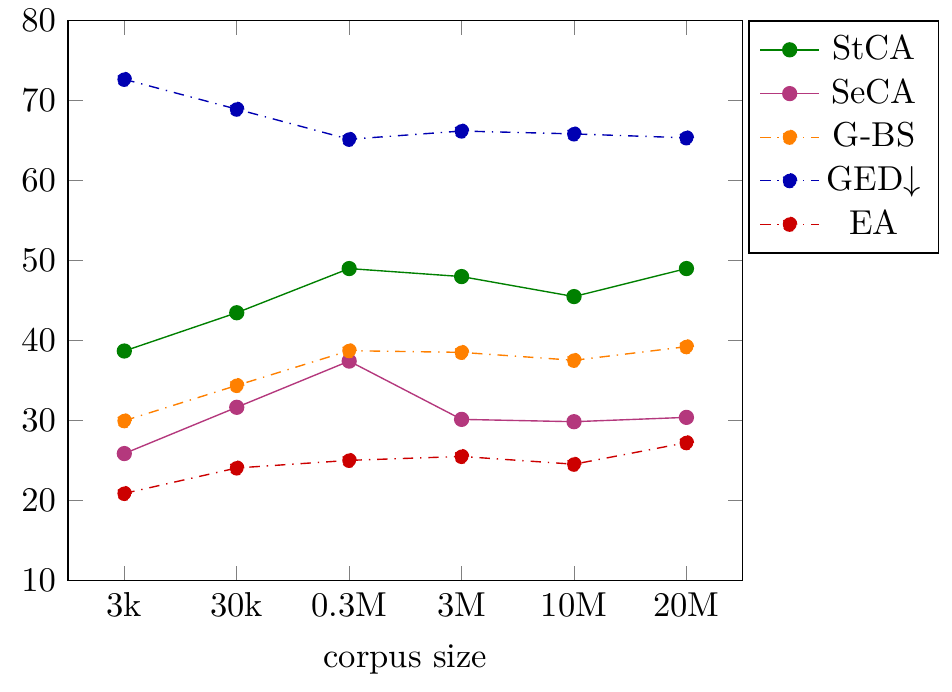}
        \caption{The results of the model pre-trained on the different difficulties of the corpus. We compared the 5 metrics for generated graphs. All the experiments use the same classifier model, reaching 88.19 on SA on the dev set.}
        \label{fig:corpusScale}
\end{figure}
Figure~\ref{fig:corpusScale} shows the results of the model pre-trained on a different scale of the corpus. 
We compare the effect of six different scales of corpus on the experiment. 
Within a certain range, the experimental results are improved by the scale of the corpus. 
However, when the corpus size exceeds a certain threshold, the marginal benefit of a larger corpus becomes increasingly diminishing, likely due to the limitations of computational resources and insufficient training on a large-scale corpus.  
Considering all factors, we select a corpus size of 0.3M as the optimal setting for our main experiments, as it yields the best results under the current conditions.

\subsection{Results on other reasoning datasets}

\begin{table}[t]
    \centering
    \resizebox{\columnwidth}{!}{
    \begin{tabular}{lcc}
    \toprule
         & w/o pre-training & w/ pre-training \\
    \hline
    CommonsenseQA   & 29.0   & 39.5      \\
    \hline
    OpenbookQA      & 34.0   & 46.0       \\
    \toprule
    \end{tabular}
    }
    \caption{The semantic accuracy of the graphs generated on CommonsenseQA and OpenbookQA by human evaluation. w/(w/o) pre-training means with(without) the step of "text-to-graph" pre-training. }
    \label{tab:OtherResults}
\end{table}
Table~\ref{tab:OtherResults} shows the results of human evaluation on CommonsenseQA(CSQA) and OpenbookQA(OBQA). 
The “text-to-graph” pre-training step improves the semantic accuracy by 10.5 on CSQA and improves the semantic accuracy by 12.0 on OBQA. 
The experimental results show that the model after “text-to-graph” pre-training is able to generate a fairly exciting explanation graph on other downstream tasks as well. 
Additionally, this serves as evidence that our methodology enhances the model's capacity for generalization.
Observing the generated graph, we find that the explanation graph generated by the model without pre-training only mechanically merely connects the question and the answer with a short path, and even generates some meaningless relations in it. More case study on these two datasets is discussed in Appendix~\ref{app:caseStudyCSQAOBQA}.

%% file: sections/related_work.tex
\section{Related Work}

\subsection{Explanation generation}

In the task of explanation generation, the model takes a piece of natural language text as input and outputs an explanation in various formats, including (a) textual highlights using a subset of the input text\cite{zaidan-etal-2007-using, lei-etal-2016-rationalizing, yu-etal-2019-rethinking, deyoung-etal-2020-eraser}, (b) natural language explanation\cite{camburu2018snli, https://doi.org/10.48550/arxiv.2010.12762, zhang-etal-2020-winowhy, inoue-etal-2021-summarize} and c) structured explanation, including semi-structured text explanation\cite{https://doi.org/10.48550/arxiv.1910.11473, https://doi.org/10.48550/arxiv.2010.03274, https://doi.org/10.48550/arxiv.2101.02235, ye-etal-2020-teaching} and structured explanation graphs\cite{jansen-etal-2018-worldtree, xie-etal-2020-worldtree, saha-etal-2021-explagraphs}. The explanation based on natural language is more expressive and easier understood by readers, but its evaluation process from the perspective of reasoning is often not standardized and rigorous\cite{wiegreffe2021teach}. Therefore, structured explanations have attracted more and more attention from researchers for they are better evaluated in terms of structure and semantics. In this paper, we choose ExplaGraphs\cite{saha-etal-2021-explagraphs} as the main experiment dataset because it is constructed based on commonsense knowledge and comes with relatively comprehensive automated evaluation metrics.

\subsection{Structured content generation from language models}
There are many kinds of works to generate structured content through language models, one of which is graph generation. Graph generation methods can be combined with various tasks, such as 
event influence graphs generation\cite{tandon-etal-2019-wiqa, madaan2020eigen}, 
temporal graphs generation\cite{https://doi.org/10.48550/arxiv.2104.00814, madaan-yang-2021-neural}, 
entailment trees generation\cite{dalvi-etal-2021-explaining}, 
knowledge graph completion\cite{li-etal-2016-commonsense, bosselut-etal-2019-comet} and 
methods for no specific semantics attached graphs generation\cite{simonovsky2018graphvae, shi2020graphaf, hwang2021symbolic}. In some other semantic parsing-related tasks, there is also the generation of structured content, such as scripts generation\cite{sakaguchi-etal-2021-proscript-partially, dalvi-etal-2019-everything, shi2022lemon} and program generation\cite{https://doi.org/10.48550/arxiv.2107.03374, liu2021tapex}. 
The graphs generated in our paper focus on all kinds of commonsense reasoning tasks. Besides, the main role of our generated graph is an explanation of the internal commonsense reasoning process based on the input.

%% file: acl2023.bbl
\begin{thebibliography}{46}
\expandafter\ifx\csname natexlab\endcsname\relax\def\natexlab#1{#1}\fi

\bibitem[{Aggarwal et~al.(2021)Aggarwal, Mandowara, Agrawal, Khandelwal,
  Singla, and Garg}]{aggarwal-etal-2021-ecqa}
Shourya Aggarwal, Divyanshu Mandowara, Vishwajeet Agrawal, Dinesh Khandelwal,
  Parag Singla, and Dinesh Garg. 2021.
\newblock \href {https://doi.org/10.18653/v1/2021.acl-long.238} {{E}xplanations
  for {C}ommonsense{QA}: {N}ew {D}ataset and {M}odels}.
\newblock In \emph{Proceedings of the 59th Annual Meeting of the Association
  for Computational Linguistics and the 11th International Joint Conference on
  Natural Language Processing (Volume 1: Long Papers)}, pages 3050--3065,
  Online. Association for Computational Linguistics.

\bibitem[{Bosselut et~al.(2019)Bosselut, Rashkin, Sap, Malaviya, Celikyilmaz,
  and Choi}]{bosselut-etal-2019-comet}
Antoine Bosselut, Hannah Rashkin, Maarten Sap, Chaitanya Malaviya, Asli
  Celikyilmaz, and Yejin Choi. 2019.
\newblock \href {https://doi.org/10.18653/v1/P19-1470} {{COMET}: Commonsense
  transformers for automatic knowledge graph construction}.
\newblock In \emph{Proceedings of the 57th Annual Meeting of the Association
  for Computational Linguistics}, pages 4762--4779, Florence, Italy.
  Association for Computational Linguistics.

\bibitem[{Camburu et~al.(2018)Camburu, Rockt{\"a}schel, Lukasiewicz, and
  Blunsom}]{camburu2018snli}
Oana-Maria Camburu, Tim Rockt{\"a}schel, Thomas Lukasiewicz, and Phil Blunsom.
  2018.
\newblock e-snli: Natural language inference with natural language
  explanations.
\newblock \emph{Advances in Neural Information Processing Systems}, 31.

\bibitem[{Chen et~al.(2021)Chen, Tworek, Jun, Yuan, Pinto, Kaplan, Edwards,
  Burda, Joseph, Brockman, Ray, Puri, Krueger, Petrov, Khlaaf, Sastry, Mishkin,
  Chan, Gray, Ryder, Pavlov, Power, Kaiser, Bavarian, Winter, Tillet, Such,
  Cummings, Plappert, Chantzis, Barnes, Herbert-Voss, Guss, Nichol, Paino,
  Tezak, Tang, Babuschkin, Balaji, Jain, Saunders, Hesse, Carr, Leike, Achiam,
  Misra, Morikawa, Radford, Knight, Brundage, Murati, Mayer, Welinder, McGrew,
  Amodei, McCandlish, Sutskever, and
  Zaremba}]{https://doi.org/10.48550/arxiv.2107.03374}
Mark Chen, Jerry Tworek, Heewoo Jun, Qiming Yuan, Henrique Ponde de~Oliveira
  Pinto, Jared Kaplan, Harri Edwards, Yuri Burda, Nicholas Joseph, Greg
  Brockman, Alex Ray, Raul Puri, Gretchen Krueger, Michael Petrov, Heidy
  Khlaaf, Girish Sastry, Pamela Mishkin, Brooke Chan, Scott Gray, Nick Ryder,
  Mikhail Pavlov, Alethea Power, Lukasz Kaiser, Mohammad Bavarian, Clemens
  Winter, Philippe Tillet, Felipe~Petroski Such, Dave Cummings, Matthias
  Plappert, Fotios Chantzis, Elizabeth Barnes, Ariel Herbert-Voss,
  William~Hebgen Guss, Alex Nichol, Alex Paino, Nikolas Tezak, Jie Tang, Igor
  Babuschkin, Suchir Balaji, Shantanu Jain, William Saunders, Christopher
  Hesse, Andrew~N. Carr, Jan Leike, Josh Achiam, Vedant Misra, Evan Morikawa,
  Alec Radford, Matthew Knight, Miles Brundage, Mira Murati, Katie Mayer, Peter
  Welinder, Bob McGrew, Dario Amodei, Sam McCandlish, Ilya Sutskever, and
  Wojciech Zaremba. 2021.
\newblock \href {https://doi.org/10.48550/ARXIV.2107.03374} {Evaluating large
  language models trained on code}.

\bibitem[{Dalvi et~al.(2021)Dalvi, Jansen, Tafjord, Xie, Smith, Pipatanangkura,
  and Clark}]{dalvi-etal-2021-explaining}
Bhavana Dalvi, Peter Jansen, Oyvind Tafjord, Zhengnan Xie, Hannah Smith,
  Leighanna Pipatanangkura, and Peter Clark. 2021.
\newblock \href {https://doi.org/10.18653/v1/2021.emnlp-main.585} {Explaining
  answers with entailment trees}.
\newblock In \emph{Proceedings of the 2021 Conference on Empirical Methods in
  Natural Language Processing}, pages 7358--7370, Online and Punta Cana,
  Dominican Republic. Association for Computational Linguistics.

\bibitem[{Dalvi et~al.(2019)Dalvi, Tandon, Bosselut, Yih, and
  Clark}]{dalvi-etal-2019-everything}
Bhavana Dalvi, Niket Tandon, Antoine Bosselut, Wen-tau Yih, and Peter Clark.
  2019.
\newblock \href {https://doi.org/10.18653/v1/D19-1457} {Everything happens for
  a reason: Discovering the purpose of actions in procedural text}.
\newblock In \emph{Proceedings of the 2019 Conference on Empirical Methods in
  Natural Language Processing and the 9th International Joint Conference on
  Natural Language Processing (EMNLP-IJCNLP)}, pages 4496--4505, Hong Kong,
  China. Association for Computational Linguistics.

\bibitem[{Devlin et~al.(2019)Devlin, Chang, Lee, and
  Toutanova}]{devlin-etal-2019-bert}
Jacob Devlin, Ming-Wei Chang, Kenton Lee, and Kristina Toutanova. 2019.
\newblock \href {https://doi.org/10.18653/v1/N19-1423} {{BERT}: Pre-training of
  deep bidirectional transformers for language understanding}.
\newblock In \emph{Proceedings of the 2019 Conference of the North {A}merican
  Chapter of the Association for Computational Linguistics: Human Language
  Technologies, Volume 1 (Long and Short Papers)}, pages 4171--4186,
  Minneapolis, Minnesota. Association for Computational Linguistics.

\bibitem[{DeYoung et~al.(2020)DeYoung, Jain, Rajani, Lehman, Xiong, Socher, and
  Wallace}]{deyoung-etal-2020-eraser}
Jay DeYoung, Sarthak Jain, Nazneen~Fatema Rajani, Eric Lehman, Caiming Xiong,
  Richard Socher, and Byron~C. Wallace. 2020.
\newblock \href {https://doi.org/10.18653/v1/2020.acl-main.408} {{ERASER}: {A}
  benchmark to evaluate rationalized {NLP} models}.
\newblock In \emph{Proceedings of the 58th Annual Meeting of the Association
  for Computational Linguistics}, pages 4443--4458, Online. Association for
  Computational Linguistics.

\bibitem[{Geva et~al.(2021)Geva, Khashabi, Segal, Khot, Roth, and
  Berant}]{https://doi.org/10.48550/arxiv.2101.02235}
Mor Geva, Daniel Khashabi, Elad Segal, Tushar Khot, Dan Roth, and Jonathan
  Berant. 2021.
\newblock \href {https://doi.org/10.48550/ARXIV.2101.02235} {Did aristotle use
  a laptop? a question answering benchmark with implicit reasoning strategies}.

\bibitem[{Hwang et~al.(2021)Hwang, Bhagavatula, Le~Bras, Da, Sakaguchi,
  Bosselut, and Choi}]{hwang2021symbolic}
Jena~D Hwang, Chandra Bhagavatula, Ronan Le~Bras, Jeff Da, Keisuke Sakaguchi,
  Antoine Bosselut, and Yejin Choi. 2021.
\newblock On symbolic and neural commonsense knowledge graphs.

\bibitem[{Inoue et~al.(2021)Inoue, Trivedi, Sinha, Balasubramanian, and
  Inui}]{inoue-etal-2021-summarize}
Naoya Inoue, Harsh Trivedi, Steven Sinha, Niranjan Balasubramanian, and Kentaro
  Inui. 2021.
\newblock \href {https://doi.org/10.18653/v1/2021.emnlp-main.490}
  {Summarize-then-answer: Generating concise explanations for multi-hop reading
  comprehension}.
\newblock In \emph{Proceedings of the 2021 Conference on Empirical Methods in
  Natural Language Processing}, pages 6064--6080, Online and Punta Cana,
  Dominican Republic. Association for Computational Linguistics.

\bibitem[{Jansen et~al.(2018)Jansen, Wainwright, Marmorstein, and
  Morrison}]{jansen-etal-2018-worldtree}
Peter Jansen, Elizabeth Wainwright, Steven Marmorstein, and Clayton Morrison.
  2018.
\newblock \href {https://aclanthology.org/L18-1433} {{W}orld{T}ree: A corpus of
  explanation graphs for elementary science questions supporting multi-hop
  inference}.
\newblock In \emph{Proceedings of the Eleventh International Conference on
  Language Resources and Evaluation ({LREC} 2018)}, Miyazaki, Japan. European
  Language Resources Association (ELRA).

\bibitem[{Jhamtani and Clark(2020)}]{https://doi.org/10.48550/arxiv.2010.03274}
Harsh Jhamtani and Peter Clark. 2020.
\newblock \href {https://doi.org/10.48550/ARXIV.2010.03274} {Learning to
  explain: Datasets and models for identifying valid reasoning chains in
  multihop question-answering}.

\bibitem[{Khot et~al.(2019)Khot, Clark, Guerquin, Jansen, and
  Sabharwal}]{https://doi.org/10.48550/arxiv.1910.11473}
Tushar Khot, Peter Clark, Michal Guerquin, Peter Jansen, and Ashish Sabharwal.
  2019.
\newblock \href {https://doi.org/10.48550/ARXIV.1910.11473} {Qasc: A dataset
  for question answering via sentence composition}.

\bibitem[{Lei et~al.(2016)Lei, Barzilay, and
  Jaakkola}]{lei-etal-2016-rationalizing}
Tao Lei, Regina Barzilay, and Tommi Jaakkola. 2016.
\newblock \href {https://doi.org/10.18653/v1/D16-1011} {Rationalizing neural
  predictions}.
\newblock In \emph{Proceedings of the 2016 Conference on Empirical Methods in
  Natural Language Processing}, pages 107--117, Austin, Texas. Association for
  Computational Linguistics.

\bibitem[{Lewis et~al.(2020)Lewis, Liu, Goyal, Ghazvininejad, Mohamed, Levy,
  Stoyanov, and Zettlemoyer}]{lewis-etal-2020-bart}
Mike Lewis, Yinhan Liu, Naman Goyal, Marjan Ghazvininejad, Abdelrahman Mohamed,
  Omer Levy, Veselin Stoyanov, and Luke Zettlemoyer. 2020.
\newblock \href {https://doi.org/10.18653/v1/2020.acl-main.703} {{BART}:
  Denoising sequence-to-sequence pre-training for natural language generation,
  translation, and comprehension}.
\newblock In \emph{Proceedings of the 58th Annual Meeting of the Association
  for Computational Linguistics}, pages 7871--7880, Online. Association for
  Computational Linguistics.

\bibitem[{Li et~al.(2016)Li, Taheri, Tu, and Gimpel}]{li-etal-2016-commonsense}
Xiang Li, Aynaz Taheri, Lifu Tu, and Kevin Gimpel. 2016.
\newblock \href {https://doi.org/10.18653/v1/P16-1137} {Commonsense knowledge
  base completion}.
\newblock In \emph{Proceedings of the 54th Annual Meeting of the Association
  for Computational Linguistics (Volume 1: Long Papers)}, pages 1445--1455,
  Berlin, Germany. Association for Computational Linguistics.

\bibitem[{Liu et~al.(2021)Liu, Chen, Guo, Ziyadi, Lin, Chen, and
  Lou}]{liu2021tapex}
Qian Liu, Bei Chen, Jiaqi Guo, Morteza Ziyadi, Zeqi Lin, Weizhu Chen, and
  Jian-Guang Lou. 2021.
\newblock Tapex: Table pre-training via learning a neural sql executor.
\newblock \emph{arXiv preprint arXiv:2107.07653}.

\bibitem[{Liu et~al.(2019)Liu, Ott, Goyal, Du, Joshi, Chen, Levy, Lewis,
  Zettlemoyer, and Stoyanov}]{roberta}
Yinhan Liu, Myle Ott, Naman Goyal, Jingfei Du, Mandar Joshi, Danqi Chen, Omer
  Levy, Mike Lewis, Luke Zettlemoyer, and Veselin Stoyanov. 2019.
\newblock \href {https://doi.org/10.48550/ARXIV.1907.11692} {Roberta: A
  robustly optimized bert pretraining approach}.

\bibitem[{Madaan et~al.(2020)Madaan, Rajagopal, Yang, Ravichander, Hovy, and
  Prabhumoye}]{madaan2020eigen}
Aman Madaan, Dheeraj Rajagopal, Yiming Yang, Abhilasha Ravichander, Eduard
  Hovy, and Shrimai Prabhumoye. 2020.
\newblock Eigen: Event influence generation using pre-trained language models.
\newblock \emph{arXiv preprint arXiv:2010.11764}.

\bibitem[{Madaan and Yang(2021)}]{madaan-yang-2021-neural}
Aman Madaan and Yiming Yang. 2021.
\newblock \href {https://doi.org/10.18653/v1/2021.naacl-main.67} {Neural
  language modeling for contextualized temporal graph generation}.
\newblock In \emph{Proceedings of the 2021 Conference of the North American
  Chapter of the Association for Computational Linguistics: Human Language
  Technologies}, pages 864--881, Online. Association for Computational
  Linguistics.

\bibitem[{Madaan et~al.(2022)Madaan, Zhou, Alon, Yang, and
  Neubig}]{madaan2022llm}
Aman Madaan, Shuyan Zhou, Uri Alon, Yiming Yang, and Graham Neubig. 2022.
\newblock Language models of code are few-shot commonsense learners.
\newblock \emph{arXiv preprint arXiv:2210.07128}.

\bibitem[{Mihaylov et~al.(2018)Mihaylov, Clark, Khot, and
  Sabharwal}]{mihaylov-etal-2018-openbookqa}
Todor Mihaylov, Peter Clark, Tushar Khot, and Ashish Sabharwal. 2018.
\newblock \href {https://doi.org/10.18653/v1/D18-1260} {Can a suit of armor
  conduct electricity? a new dataset for open book question answering}.
\newblock In \emph{Proceedings of the 2018 Conference on Empirical Methods in
  Natural Language Processing}, pages 2381--2391, Brussels, Belgium.
  Association for Computational Linguistics.

\bibitem[{Ott et~al.(2019)Ott, Edunov, Baevski, Fan, Gross, Ng, Grangier, and
  Auli}]{ott-etal-2019-fairseq}
Myle Ott, Sergey Edunov, Alexei Baevski, Angela Fan, Sam Gross, Nathan Ng,
  David Grangier, and Michael Auli. 2019.
\newblock \href {https://doi.org/10.18653/v1/N19-4009} {fairseq: A fast,
  extensible toolkit for sequence modeling}.
\newblock In \emph{Proceedings of the 2019 Conference of the North {A}merican
  Chapter of the Association for Computational Linguistics (Demonstrations)},
  pages 48--53, Minneapolis, Minnesota. Association for Computational
  Linguistics.

\bibitem[{Raffel et~al.(2020)Raffel, Shazeer, Roberts, Lee, Narang, Matena,
  Zhou, Li, and Liu}]{t5}
Colin Raffel, Noam Shazeer, Adam Roberts, Katherine Lee, Sharan Narang, Michael
  Matena, Yanqi Zhou, Wei Li, and Peter~J. Liu. 2020.
\newblock \href {http://jmlr.org/papers/v21/20-074.html} {Exploring the limits
  of transfer learning with a unified text-to-text transformer}.
\newblock \emph{Journal of Machine Learning Research}, 21(140):1--67.

\bibitem[{Rajagopal et~al.(2021)Rajagopal, Madaan, Tandon, Yang, Prabhumoye,
  Ravichander, Clark, and Hovy}]{https://doi.org/10.48550/arxiv.2104.00814}
Dheeraj Rajagopal, Aman Madaan, Niket Tandon, Yiming Yang, Shrimai Prabhumoye,
  Abhilasha Ravichander, Peter Clark, and Eduard Hovy. 2021.
\newblock \href {https://doi.org/10.48550/ARXIV.2104.00814} {Curie: An
  iterative querying approach for reasoning about situations}.

\bibitem[{Saha et~al.(2020)Saha, Ghosh, Srivastava, and
  Bansal}]{saha-etal-2020-prover}
Swarnadeep Saha, Sayan Ghosh, Shashank Srivastava, and Mohit Bansal. 2020.
\newblock \href {https://doi.org/10.18653/v1/2020.emnlp-main.9} {{PR}over:
  Proof generation for interpretable reasoning over rules}.
\newblock In \emph{Proceedings of the 2020 Conference on Empirical Methods in
  Natural Language Processing (EMNLP)}, pages 122--136, Online. Association for
  Computational Linguistics.

\bibitem[{Saha et~al.(2022)Saha, Yadav, and
  Bansal}]{saha-etal-2022-explanation}
Swarnadeep Saha, Prateek Yadav, and Mohit Bansal. 2022.
\newblock \href {https://doi.org/10.18653/v1/2022.acl-long.85} {Explanation
  graph generation via pre-trained language models: An empirical study with
  contrastive learning}.
\newblock In \emph{Proceedings of the 60th Annual Meeting of the Association
  for Computational Linguistics (Volume 1: Long Papers)}, pages 1190--1208,
  Dublin, Ireland. Association for Computational Linguistics.

\bibitem[{Saha et~al.(2021)Saha, Yadav, Bauer, and
  Bansal}]{saha-etal-2021-explagraphs}
Swarnadeep Saha, Prateek Yadav, Lisa Bauer, and Mohit Bansal. 2021.
\newblock \href {https://doi.org/10.18653/v1/2021.emnlp-main.609}
  {{E}xpla{G}raphs: An explanation graph generation task for structured
  commonsense reasoning}.
\newblock In \emph{Proceedings of the 2021 Conference on Empirical Methods in
  Natural Language Processing}, pages 7716--7740, Online and Punta Cana,
  Dominican Republic. Association for Computational Linguistics.

\bibitem[{Sakaguchi et~al.(2021)Sakaguchi, Bhagavatula, Le~Bras, Tandon, Clark,
  and Choi}]{sakaguchi-etal-2021-proscript-partially}
Keisuke Sakaguchi, Chandra Bhagavatula, Ronan Le~Bras, Niket Tandon, Peter
  Clark, and Yejin Choi. 2021.
\newblock \href {https://doi.org/10.18653/v1/2021.findings-emnlp.184}
  {pro{S}cript: Partially ordered scripts generation}.
\newblock In \emph{Findings of the Association for Computational Linguistics:
  EMNLP 2021}, pages 2138--2149, Punta Cana, Dominican Republic. Association
  for Computational Linguistics.

\bibitem[{Saxena et~al.(2022)Saxena, Kochsiek, and
  Gemulla}]{saxena-etal-2022-kgt5}
Apoorv Saxena, Adrian Kochsiek, and Rainer Gemulla. 2022.
\newblock \href {https://doi.org/10.18653/v1/2022.acl-long.201}
  {Sequence-to-sequence knowledge graph completion and question answering}.
\newblock In \emph{Proceedings of the 60th Annual Meeting of the Association
  for Computational Linguistics (Volume 1: Long Papers)}, pages 2814--2828,
  Dublin, Ireland. Association for Computational Linguistics.

\bibitem[{Shi et~al.(2020)Shi, Xu, Zhu, Zhang, Zhang, and
  Tang}]{shi2020graphaf}
Chence Shi, Minkai Xu, Zhaocheng Zhu, Weinan Zhang, Ming Zhang, and Jian Tang.
  2020.
\newblock Graphaf: a flow-based autoregressive model for molecular graph
  generation.
\newblock \emph{arXiv preprint arXiv:2001.09382}.

\bibitem[{Shi et~al.(2022)Shi, Liu, Chen, Zhang, Liu, and Lou}]{shi2022lemon}
Qi~Shi, Qian Liu, Bei Chen, Yu~Zhang, Ting Liu, and Jian-Guang Lou. 2022.
\newblock Lemon: Language-based environment manipulation via execution-guided
  pre-training.
\newblock \emph{arXiv preprint arXiv:2201.08081}.

\bibitem[{Simonovsky and Komodakis(2018)}]{simonovsky2018graphvae}
Martin Simonovsky and Nikos Komodakis. 2018.
\newblock Graphvae: Towards generation of small graphs using variational
  autoencoders.
\newblock In \emph{International conference on artificial neural networks},
  pages 412--422. Springer.

\bibitem[{Speer et~al.(2017)Speer, Chin, and Havasi}]{speer2017conceptnet}
Robyn Speer, Joshua Chin, and Catherine Havasi. 2017.
\newblock Conceptnet 5.5: An open multilingual graph of general knowledge.

\bibitem[{Talmor et~al.(2019)Talmor, Herzig, Lourie, and
  Berant}]{talmor-etal-2019-commonsenseqa}
Alon Talmor, Jonathan Herzig, Nicholas Lourie, and Jonathan Berant. 2019.
\newblock \href {https://doi.org/10.18653/v1/N19-1421} {{C}ommonsense{QA}: A
  question answering challenge targeting commonsense knowledge}.
\newblock In \emph{Proceedings of the 2019 Conference of the North {A}merican
  Chapter of the Association for Computational Linguistics: Human Language
  Technologies, Volume 1 (Long and Short Papers)}, pages 4149--4158,
  Minneapolis, Minnesota. Association for Computational Linguistics.

\bibitem[{Tandon et~al.(2019)Tandon, Dalvi, Sakaguchi, Clark, and
  Bosselut}]{tandon-etal-2019-wiqa}
Niket Tandon, Bhavana Dalvi, Keisuke Sakaguchi, Peter Clark, and Antoine
  Bosselut. 2019.
\newblock \href {https://doi.org/10.18653/v1/D19-1629} {{WIQA}: A dataset for
  {``}what if...{''} reasoning over procedural text}.
\newblock In \emph{Proceedings of the 2019 Conference on Empirical Methods in
  Natural Language Processing and the 9th International Joint Conference on
  Natural Language Processing (EMNLP-IJCNLP)}, pages 6076--6085, Hong Kong,
  China. Association for Computational Linguistics.

\bibitem[{Wiegreffe and Marasovi{\'c}(2021)}]{wiegreffe2021teach}
Sarah Wiegreffe and Ana Marasovi{\'c}. 2021.
\newblock Teach me to explain: A review of datasets for explainable nlp.
\newblock \emph{arXiv preprint arXiv:2102.12060}.

\bibitem[{Wiegreffe et~al.(2020)Wiegreffe, Marasović, and
  Smith}]{https://doi.org/10.48550/arxiv.2010.12762}
Sarah Wiegreffe, Ana Marasović, and Noah~A. Smith. 2020.
\newblock \href {https://doi.org/10.48550/ARXIV.2010.12762} {Measuring
  association between labels and free-text rationales}.

\bibitem[{Xie et~al.(2020)Xie, Thiem, Martin, Wainwright, Marmorstein, and
  Jansen}]{xie-etal-2020-worldtree}
Zhengnan Xie, Sebastian Thiem, Jaycie Martin, Elizabeth Wainwright, Steven
  Marmorstein, and Peter Jansen. 2020.
\newblock \href {https://aclanthology.org/2020.lrec-1.671} {{W}orld{T}ree v2: A
  corpus of science-domain structured explanations and inference patterns
  supporting multi-hop inference}.
\newblock In \emph{Proceedings of the Twelfth Language Resources and Evaluation
  Conference}, pages 5456--5473, Marseille, France. European Language Resources
  Association.

\bibitem[{Yang et~al.(2018)Yang, Qi, Zhang, Bengio, Cohen, Salakhutdinov, and
  Manning}]{yang-etal-2018-hotpotqa}
Zhilin Yang, Peng Qi, Saizheng Zhang, Yoshua Bengio, William Cohen, Ruslan
  Salakhutdinov, and Christopher~D. Manning. 2018.
\newblock \href {https://doi.org/10.18653/v1/D18-1259} {{H}otpot{QA}: A dataset
  for diverse, explainable multi-hop question answering}.
\newblock In \emph{Proceedings of the 2018 Conference on Empirical Methods in
  Natural Language Processing}, pages 2369--2380, Brussels, Belgium.
  Association for Computational Linguistics.

\bibitem[{Ye et~al.(2020)Ye, Huang, Boschee, and Ren}]{ye-etal-2020-teaching}
Qinyuan Ye, Xiao Huang, Elizabeth Boschee, and Xiang Ren. 2020.
\newblock \href {https://doi.org/10.18653/v1/2020.findings-emnlp.145} {Teaching
  machine comprehension with compositional explanations}.
\newblock In \emph{Findings of the Association for Computational Linguistics:
  EMNLP 2020}, pages 1599--1615, Online. Association for Computational
  Linguistics.

\bibitem[{Yu et~al.(2019)Yu, Chang, Zhang, and
  Jaakkola}]{yu-etal-2019-rethinking}
Mo~Yu, Shiyu Chang, Yang Zhang, and Tommi Jaakkola. 2019.
\newblock \href {https://doi.org/10.18653/v1/D19-1420} {Rethinking cooperative
  rationalization: Introspective extraction and complement control}.
\newblock In \emph{Proceedings of the 2019 Conference on Empirical Methods in
  Natural Language Processing and the 9th International Joint Conference on
  Natural Language Processing (EMNLP-IJCNLP)}, pages 4094--4103, Hong Kong,
  China. Association for Computational Linguistics.

\bibitem[{Zaidan et~al.(2007)Zaidan, Eisner, and
  Piatko}]{zaidan-etal-2007-using}
Omar Zaidan, Jason Eisner, and Christine Piatko. 2007.
\newblock \href {https://aclanthology.org/N07-1033} {Using {``}annotator
  rationales{''} to improve machine learning for text categorization}.
\newblock In \emph{Human Language Technologies 2007: The Conference of the
  North {A}merican Chapter of the Association for Computational Linguistics;
  Proceedings of the Main Conference}, pages 260--267, Rochester, New York.
  Association for Computational Linguistics.

\bibitem[{Zhang et~al.(2020)Zhang, Zhao, and Song}]{zhang-etal-2020-winowhy}
Hongming Zhang, Xinran Zhao, and Yangqiu Song. 2020.
\newblock \href {https://doi.org/10.18653/v1/2020.acl-main.508} {{W}ino{W}hy: A
  deep diagnosis of essential commonsense knowledge for answering {W}inograd
  schema challenge}.
\newblock In \emph{Proceedings of the 58th Annual Meeting of the Association
  for Computational Linguistics}, pages 5736--5745, Online. Association for
  Computational Linguistics.

\bibitem[{Zhang et~al.(2019)Zhang, Kishore, Wu, Weinberger, and
  Artzi}]{https://doi.org/10.48550/arxiv.1904.09675}
Tianyi Zhang, Varsha Kishore, Felix Wu, Kilian~Q. Weinberger, and Yoav Artzi.
  2019.
\newblock \href {https://doi.org/10.48550/ARXIV.1904.09675} {Bertscore:
  Evaluating text generation with bert}.

\end{thebibliography}
